\documentclass[runningheads]{llncs}

\usepackage{eccv}

\usepackage{eccvabbrv}

\usepackage{graphicx}
\usepackage{booktabs}

\usepackage[accsupp]{axessibility}  

\usepackage{hyperref}

\usepackage{orcidlink}

\newcommand{\myparagraph}[1]{\noindent\textbf{#1\ }}

\newcommand\redsout{\bgroup\markoverwith{\textcolor{red}{\rule[0.5ex]{2pt}{1pt}}}\ULon}

\newcommand{\sieveswap}{Sieve\,\&\,Swap\xspace}%
\newcommand{\sieveswapnospace}{\sieveswap}%
\newcommand{\cmark}{\ding{51}}%
\newcommand{\xmark}{\ding{55}}%
\newcommand{\specialcell}[2][c]{%
  \begin{tabular}[#1]{@{}c@{}}#2\end{tabular}}

\newcommand{\gray}[1]{{\color{gray!100} #1}}

\newcommand{\highlight}[1]{%
  \colorbox{blue!10}{$\displaystyle#1$}}

\usepackage{amsmath}	
\usepackage{amssymb}	
\usepackage{microtype}
\usepackage{epsfig}
\usepackage{caption}
\usepackage{float}
\usepackage{placeins}
\usepackage{color, colortbl}
\usepackage{stfloats}
\usepackage{enumitem}
\usepackage{tabularx}
\usepackage{xstring}
\usepackage{multirow}
\usepackage{xspace}
\usepackage{url}
\usepackage{subcaption}
\usepackage[hang,flushmargin]{footmisc}
\usepackage{pifont}
\usepackage{svg}
\usepackage{soul}
\usepackage{parskip}
\usepackage{arydshln}
\usepackage[normalem]{ulem}
\usepackage{adjustbox}

\begin{document}

\newcolumntype{a}{>{\columncolor{SpringGreen}}c}

\title{Efficient Pre-training for Localized Instruction Generation of Procedural Videos} 

\titlerunning{\sieveswap: Efficient Pre-training of Procedural Videos}

\author{Anil Batra\inst{\spadesuit}\orcidlink{0009-0005-1919-9712} \and
Davide Moltisanti\inst{\blacklozenge}\orcidlink{0000-0003-4265-8882} \and
Laura Sevilla-Lara\inst{\spadesuit}\orcidlink{0000-0001-8276-0094} \and
Marcus Rohrbach\inst{\bigstar}\orcidlink{0000-0001-5908-7751} \and
Frank Keller\inst{\spadesuit}\orcidlink{0000-0002-8242-4362}
}
\authorrunning{Anil Batra et al.}

\institute{$^\spadesuit$University of Edinburgh \qquad $^\blacklozenge$University of Bath \qquad $^\bigstar$TU Darmstadt \& hessian.AI\\
\email{a.k.batra@ed.ac.uk}}

\maketitle

\begin{abstract}
Procedural videos, exemplified by recipe demonstrations, are instrumental in conveying step-by-step instructions. However, understanding such videos is challenging as it involves the precise localization of steps and the generation of textual instructions. Manually annotating steps and writing instructions is costly, which limits the size of current datasets and hinders effective learning. Leveraging large but noisy video-transcript datasets for pre-training can boost performance but demands significant computational resources. Furthermore, transcripts contain irrelevant content and differ in style from human-written instructions. To mitigate these issues, we propose a novel technique, \sieveswap, to automatically generate high-quality training data for the recipe domain: (i)~Sieve: filters irrelevant transcripts and (ii)~Swap: acquires high-quality text by replacing transcripts with human-written instruction from a text-only recipe dataset. The resulting dataset is three orders of magnitude smaller than current web-scale datasets but enables efficient training of large-scale models. Alongside \sieveswap, we propose Procedure Transformer (ProcX), a model for end-to-end step localization and instruction generation for procedural videos. When pre-trained on our curated dataset, this model achieves state-of-the-art performance on YouCook2 and Tasty while using a fraction of the training data. We have released code and dataset.\footnote{\url{https://github.com/anilbatra2185/sns_procx}}
\end{abstract}
\section{Introduction}
\label{sec:intro}
\begin{figure}
    \centering
    \begin{subfigure}{.65\textwidth}
      \centering
        \includegraphics[width=0.99\linewidth]{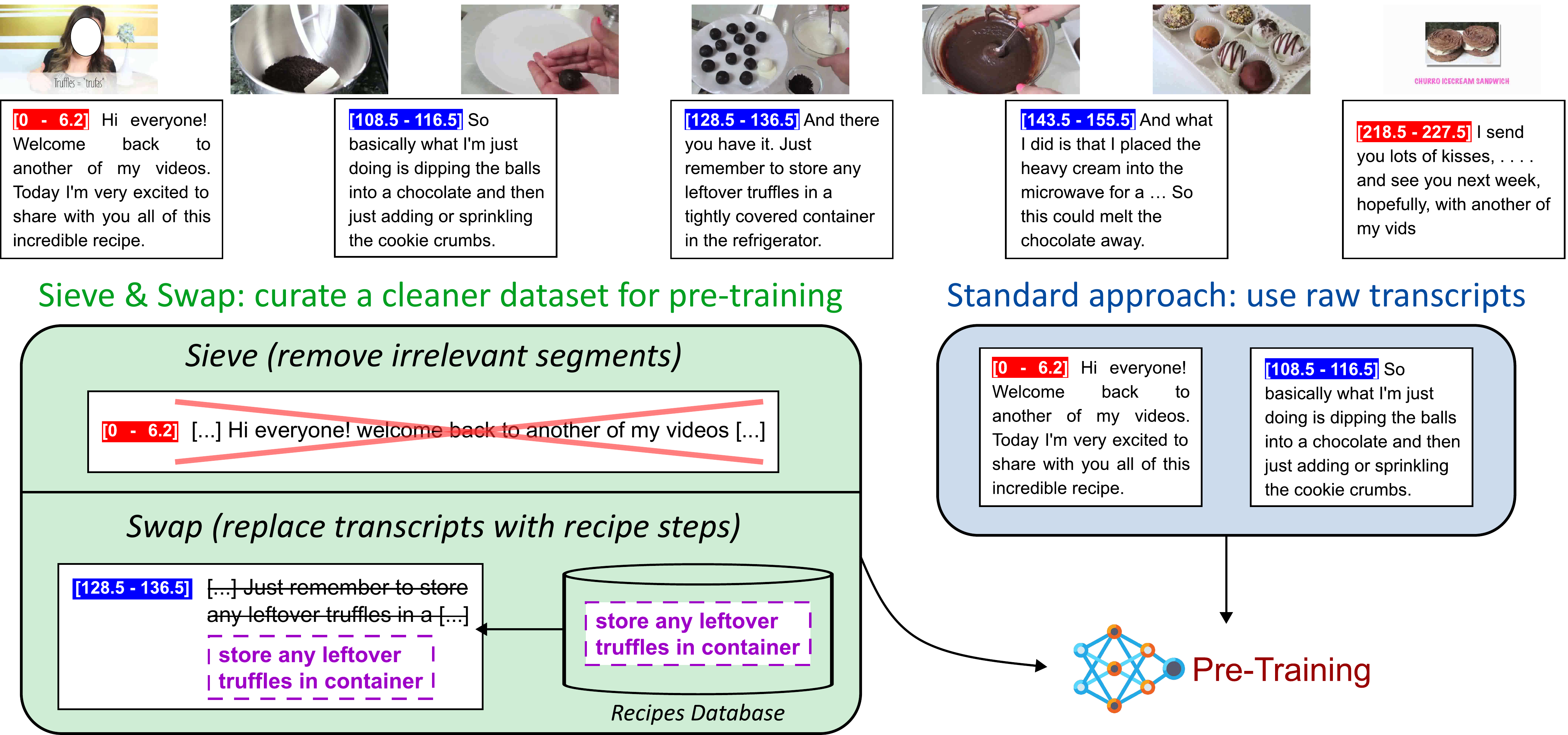}
        \caption{}
        \label{fig:teaser_sievenswap}
    \end{subfigure}%
    \hfill
    \begin{subfigure}{.32\textwidth}
      \centering
        \includegraphics[width=0.99\linewidth]{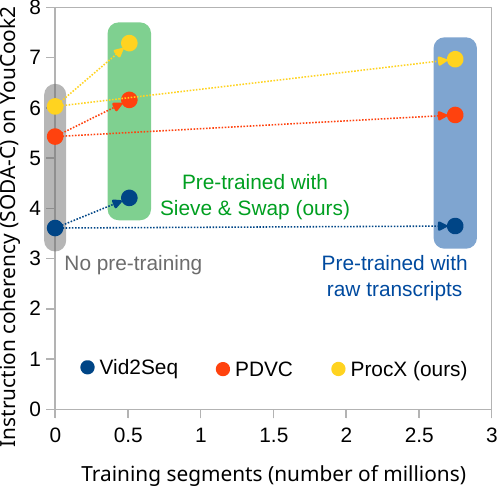}
        \caption{}
        \label{fig:teaser_results}
    \end{subfigure}
    \captionsetup{font=scriptsize}
    \caption{
        \textbf{(a)} 
        \sieveswap at a glance. We first remove irrelevant video ASR (Automatic Speech Recognition) segments, then substitute the raw transcripts with recipe steps retrieved from a recipe database. The resulting dataset is used for pre-training. Previous work uses raw and noisier transcripts, thus requiring larger amounts of data for effective pre-training.
        \textbf{(b)} With \sieveswap we generate a smaller but better pre-training dataset. Compared to using a larger number of raw segments (right), models achieve higher performance with a fraction of the data (middle). Note how the improvement from no pre-training (left) is steeper with \sieveswapnospace. The plotted metric is the generated instruction coherency metric (SODA-C \cite{fujita2020soda}), models were fine-tuned and tested on YouCook2 \cite{ZhXuCoCVPR18}.}
        \label{fig:teaser}
\end{figure}
Instructional videos are abundantly available online and constitute a valuable resource for humans to acquire procedural knowledge, aiding in tasks like recipe preparation. However, the task of understanding the sequence of actions in instructional videos is challenging for current AI models. It involves two distinct tasks: (i)~the temporal localization of steps in the video and (ii)~the generation of a textual instruction for each step. There are a number of existing datasets for procedural learning, including COIN \cite{tang2019coin} and CrossTask \cite{zhukov2019cross}, which are annotated with fixed, template-based labels (e.g.,~\textit{pour-water}). These labels, however, fail to describe the varied details of instructions in the videos (e.g.,~attributes and fine-grained objects). Datasets annotated with human-written instructions exist in the recipe domain, in particular YouCook2 \cite{ZhXuCoCVPR18} and Tasty \cite{sener2019zero}. However, due to the substantial annotation cost these datasets are limited in size, which poses a risk of over-fitting for large models. To achieve good performance with current neural architectures, large amounts of training data are required \cite{cheng2023vindlu}, and researchers have addressed this challenge by augmenting datasets or adopting strategies such as self-supervised training or pre-training with noisy web-scale datasets. Drawing inspiration from substantial advancements in caption generation achieved with datasets such as HowTo100M \cite{miech2019howto100m}, Yang \etal \cite{yang2023vid2seq} employ noisy video transcripts (automatically transcribed speech) as a form of weak supervision during pre-training to generate segment boundaries and instructions. Leveraging noisy data is beneficial, however this method trains on one million videos, requiring at least 12 TB of storage in low quality.

Another limitation of this strategy is sub-optimal multimodal learning due to irrelevant content and stylistic differences between transcripts and written language. For example, transcripts often include content like greetings and advertisements, which do not form part of the instructions that video segments are annotated with. Studies by Chafe \etal \cite{chafe1987relation} and Einhorn \etal \cite{einhorn1978oral} highlight the distinct attributes of written and spoken language: written language is more concise and varied, while spoken language tends to be more protracted and repetitive. This creates a substantial domain gap. In this work, we propose \sieveswapnospace, a technique that fuses an instructional video dataset and a recipe text dataset, resulting in a smaller multimodal dataset for effective procedure learning. Figure~\ref{fig:teaser_sievenswap} summarizes our approach and contrasts it with prior methods. Specifically, we combine a subset of HowTo100M \cite{miech2019howto100m} of instructional cooking videos with the RecipeNLG \cite{bien2020recipenlg} collection of text-only cooking recipes. We develop a novel method to retrieve relevant and high-quality reference descriptions for procedural learning, which involves extracting human-written instructional sentences from RecipeNLG. We employ sentence embeddings to filter irrelevant sentences from video transcripts based on text similarity, and replace the relevant transcript sentences with instructional sentences from RecipeNLG. The resulting \sieveswap dataset is characterized by less noise, providing high-quality pre-training data for the task of human-style instruction generation. The \sieveswap dataset is smaller by three orders of magnitude ($\approx$48K videos) than the pre-training datasets employed in previous studies ($\approx$15M videos) \cite{Luo2020UniVL, yang2023vid2seq}, but contains both segments with temporal boundaries and human-written instructions. Despite the smaller scale, models pre-trained on the \sieveswap dataset outperform the same models pre-trained on raw transcripts, as we show in Figure~\ref{fig:teaser_results}.

Our second contribution is Procedure Transformer ({\em ProcX}), a model that combines pre-training and fine-tuning to localize steps in videos and generate textual instructions. ProcX is designed for effective pre-training on our \sieveswap dataset. It builds on the PDVC model \cite{wang2021end, batra2022closer}, but introduces a number of innovations. First, we note that set-based transformers (DETR \cite{carion2020end}) with deformable attention \cite{zhudeformable} achieved significant improvements for temporal action detection \cite{liu2022end, shi2022react} and dense video captioning \cite{wang2021end}. However, the naive scaling of prior models, such as PDVC, is non-trivial. For instance, deformable attention struggles to effectively utilize multi-scale temporal features, and LSTM-based text generation lacks parallelism and is difficult to train. To address these issues, ProcX combines key-aware deformable attention with a contrastive transformer for text generation, using an IoU-aware confidence score during inference. We validate each of these improvements in an ablation study and show that ProcX achieves state-of-the-art performance for procedure learning (see Figure~\ref{fig:teaser_results}). 

To summarize, our contributions are:
\begin{enumerate}
    \item We propose a novel approach, \sieveswap, to generate a dataset for pre-training procedural learning models automatically. Our approach reduces the domain gap between pre-training text (transcribed speech) and target text (written instructions).
    \item Using this approach, we create the \sieveswap dataset for pre-training models for procedural video understanding in the cooking domain. This dataset is three orders of magnitude smaller than previous datasets, allowing efficient pre-training of instruction localization and generation models.
    \item We propose Procedure Transformer ({\em ProcX}), an improved instruction localization and description model, that we pre-train on the \sieveswap dataset to achieve state-of-the-art performance on YouCook2 and Tasty.
\end{enumerate}
\section{Related Work}
\label{sec:related}
Procedural learning is a challenging task in the field of video understanding, as it focuses on acquiring procedural knowledge from long untrimmed videos. Motivated by dense video captioning in open domain videos \cite{krishna2017dense}, Zhou \etal \cite{ZhXuCoCVPR18} were the first to introduce a dataset and task in the cooking domain to learn procedures from videos. Annotation styles for procedural learning include fixed, template-based labels (e.g., COIN \cite{tang2019coin}, CrossTask \cite{zhukov2019cross}) and human-written instructions (e.g., YouCook2 \cite{ZhXuCoCVPR18}, Tasty \cite{sener2019zero}). While the former uses a limited set of labels (e.g., \textit{pour-water}), the latter offers detailed descriptions but is expensive to annotate manually, resulting in smaller datasets (e.g., YouCook2 \cite{ZhXuCoCVPR18} with $\approx$1.3K videos). We focus on procedural learning by temporally localizing the events and generating detailed textual instructions. We operate in the cooking domain as it offers a diversity of both natural language and visual activities. Due to task complexity, prior work addresses the problem with two-stage training, where events are first localized and captions are later generated. Wang \etal \cite{wang2021end} introduced single-stage model (PDVC) to jointly identify temporal boundaries and generate procedural instructions. 

Recently, Yang \etal \cite{yang2023vid2seq} introduced Vid2Seq, a new large model that leverages web-scale datasets (HowTo100M~\cite{miech2019howto100m}, YT-Temporal-1B \cite{zellers2022merlot}) to enhance performance through pre-training and finetuning. Unlike conventional methods, Yang \etal \cite{yang2023vid2seq} employed speech transcripts as a form of weak supervision. The transcribed speech is given in input during inference as well, which greatly boosts Vid2Seq's performanceowever, utilizing transcript sentences converts the problem into text-to-text translation, which makes the model less effective in utilizing the video modality. Furthermore, in many videos the transcribed speech may be unavailable and the model needs to rely on the visual modality alone. The Tasty dataset \cite{sener2019zero}, derived from a web database, serves as an example of a dataset designed for this purpose. In contrast, we focus on models given only video input during training and inference time and reduce the size of the pre-training dataset (48K videos), requiring far less compute. Despite the limited computation budget we achieve state-of-the-art results thanks to our method that compiles a smaller and cleaner pre-training dataset and a more efficient transformer model.

External textual databases~\cite{wikiurl, koupaee2018wikihow} have been explored in prior work~\cite{lin2022learning, zhou2023procedure} to curate pre-training data for procedural videos. However, this work focuses on video-segment representation learning, while we focus on generating fine-grained textual instructions along with temporal localization of key steps. Zhou~\etal~\cite{zhou2023procedure} employ visual content to align video segments with the nearest step from a textual database using multi-modal embeddings, without relying on ASR text. Conversely, Lin~\etal~\cite{lin2022learning} match each ASR text to the nearest step in a textual database to learn video-segment representation. Unlike prior work~\cite{lin2022learning, zhou2023procedure}, we sieve noisy ASR segments and replace conversational speech with human-written text without visual training. Our method uses five times less text than~\cite{lin2022learning}, which uses all ASR-transcribed text. Moreover, the text-replacement approach is crucial for human-style instruction generation.

\section{The Task: Localized Instruction Generation}
 In this work, we are given an untrimmed procedural video showing a sequence of complex instructions that are necessary to complete a goal (e.g., a person following a cooking recipe). 
Each instruction is labeled with the start and end times delimiting the instruction temporally, as well as with a fine-grained textual description. Our task is to output the start/end times of each instruction, together with its textual description, given only the untrimmed video as input. This task has been termed ``Procedure Segmentation and Summarization'' (PSS) in~\cite{batra2022closer} and overlaps with Dense Video Captioning (DVC). In both cases the goal is to detect and describe activities in a video; however, key distinctions are that DVC doesn't involve instructions and that it allows overlapping segments. In our task, instructions are sequential parts of a single procedure (e.g., a recipe), and the focus is to provide a textual sequence summarizing the procedure. In fact, popular datasets for DVC such as ActivityNet Captions~\cite{krishna2017dense} collect non-procedural videos, i.e., captions describe what happens in the video, but they do not form the textual sequence of a procedure, which is the focus of our task. 

One of the main challenges in localized instruction generation is finding an adequate training dataset since we cannot rely on the large ActivityNet Captions~\cite{krishna2017dense}. Other resources provide high-quality annotation of procedural videos (e.g., YouCook2 \cite{ZhXuCoCVPR18}), however, they are limited in scale and can only be used for fine-tuning or testing. The most common approach to tackle this problem is to pre-train on vast but noisy procedural datasets such a HowTo100M~\cite{miech2019howto100m}, which incurs large computation and storage costs. We instead propose to automatically curate a significantly smaller but cleaner dataset with our \sieveswap method, which we detail next. In Section~\ref{sec:procx} we also introduce a more efficient Transformer for this task.

\section{\sieveswap}
In this section we develop a framework to curate a dataset for efficient pre-training of instructional videos with an automatic pipeline.
\subsection{Objective}
Let a corpus $\mathcal{V}$ denote the set of untrimmed procedural videos with corresponding ASR (automatic speech recognition) transcripts. Our goal is to sieve out irrelevant transcripts and swap them with human-generated text sourced from an instructional knowledge base $\mathcal{R}$, while ensuring temporal alignment with the videos. In this case, $\mathcal{R}$ is a collection of text-only recipes, each comprising sequential steps. Formally, for each video $\mathbf{v}$ we obtain the transcribed speech with start $t^s$ and end timestamps $t^e$ recognized by an ASR system:
\begin{align}
    \mathbf{v} &= (\mathcal{A}, T^v) \label{eq:video_tuple}\\
    \mathcal{A} &= \{(a_j, t^s_j, t^e_j)\}_{j \leq M} \label{eq:asr_tuple}
\end{align}
where $a$ is the automatically transcribed caption, $t^s$/$t^e$ are the start/end times of the transcript segment, $T^v$ is the video title and $M$ is the number of ASR segments in the video. Our goal is to remove irrelevant transcript segments and substitute filtered transcripts with text from $\mathcal{R}$ to create an instruction sequence:
\begin{align}
    \sieveswap(\mathbf{v}) &= (\mathcal{A}^{\sieveswap}, T^v) \\
    \mathcal{A}^{\sieveswap} &= \{(q_i, t^s_i, t^e_i)\}_{i \leq N}    
\end{align}
where $q_i$ is the recipe step retrieved from the recipe dataset $\mathcal{R}$ describing the instruction in the corresponding video segment, and $N$ is the number of filtered segments in the video. The number of instructional steps is less than the number of ASR transcript segments, i.e., $N \ll M$. While we replace the transcript text we keep the timestamps of the transcripts. The generated corpus serves as pre-training data for procedure learning, providing language supervision in the form of human-written instruction text rather than noisy ASR transcripts.
\subsection{Dataset Creation}
\label{sec:dataset_creation}
\begin{figure*}[t]
    \centering
    \includegraphics[width=0.99\linewidth]{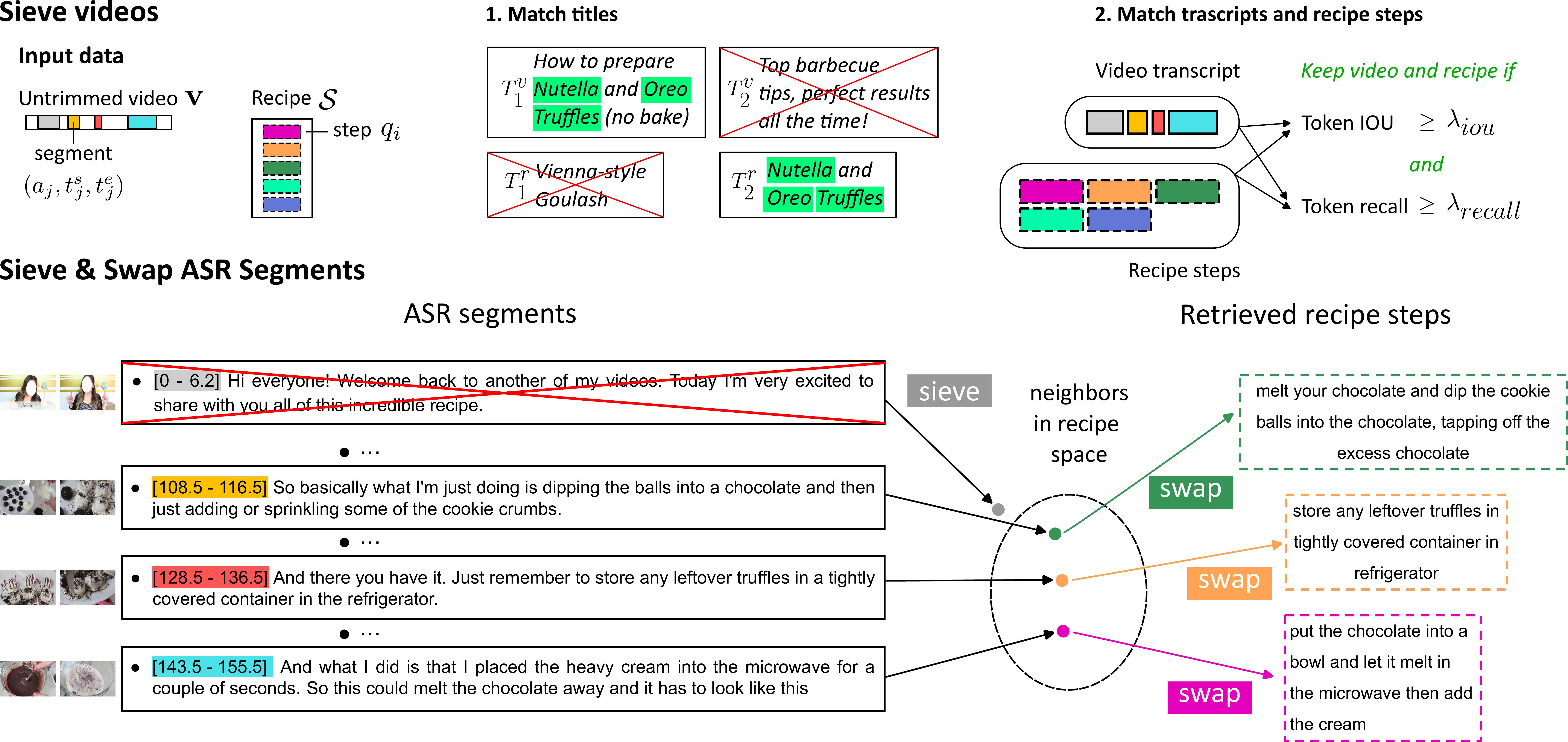}
    \captionsetup{font=scriptsize}
    \caption{\textbf{\sieveswap:} A technique to sieve and swap ASR transcripts with human written instructions. Top: we first sieve videos and recipes matching their titles and text content. Bottom: for each ASR transcript in each filtered video, we obtain a text embedding and look for the nearest neighbor among the text embeddings of each recipe step from the filtered recipes. If the retrieved neighbor is too far away (gray dots) we consider the transcript segment irrelevant and discard it. Otherwise (colored dots) we replace it with the recipe step corresponding to the retrieved neighbor.}
    \label{fig:data_create_stage2}
\end{figure*}
\subsubsection{Source Datasets.} \label{sec:dataset_creation_1}
We utilize all cooking videos from HowTo100M ($\approx$338K) and all the cooking recipes with steps from RecipesNLG~\cite{bien2020recipenlg} ($\approx$2M). HowTo100M is the largest dataset for pre-training video models, while RecipesNLG is collected to generate structured text sequences. We keep videos with a maximum duration of 10 minutes and require a minimum of five videos per cooking category (e.g.,~`make a bean salad') obtained from video metadata. This set of videos is the corpus $\mathcal{V}$ with $\approx$241K videos paired with a title containing a high-level goal or a cooking recipe title. We first sieve irrelevant videos and recipes by pairing them based on their title and text content; we then apply \sieveswap on ASR segments to discard irrelevant ASR text (e.g., greetings) and refine segments by substituting ASR transcripts with steps from the recipes database.

\subsubsection{Sieving Videos.} \label{sec:dataset_creation_2}
The source dataset contains a large amount of noisy videos and requires large computational resources. We filter the video dataset by pairing each video $\mathbf{v}$ with a set of textual recipes based on (i) word overlap between video and recipe titles and (ii) word overlap between ASR Transcript and recipe instructions. Let $\mathcal{R} = \{(\mathcal{S}_i, T^r_i)\}$ be the textual recipe dataset, where $\mathcal{S}_i = \{q_1, \dots, q_H\}$ is the set of procedure steps in a recipe and $T^r_i$ is the recipe title. We assume that the titles $T^v$ and $T^r$ have enough semantic information to map a video $\mathbf{v}$ to a set of recipes in $\mathcal{R}$. With this we refine both $\mathcal{V}$ and $\mathcal{R}$: 
\begin{equation}
    \mathcal{V}^1 = \{\mathbf{v} = (\mathcal{A}, T^v) \ | \ \exists \{ T^v \cap T^r\} \neq \emptyset; \ \forall \mathbf{v} \in \mathcal{V} ,\ \forall T^r \in \mathcal{R} \}
\end{equation}
\noindent To calculate the overlap between titles  ($T^v \cap T^r$), we consider only content words (nouns, verbs, adverbs, adjectives) and ignore generic recipe words such as `make', `prepare', and `bake'. Videos without a recipe mapping are removed, and recipes that are not paired to any video are likewise removed. We denote the new set of recipes with $\mathcal{R}^1$. The size of $\mathcal{V}^1$ at this stage is $\approx$110K videos.

We filter $\mathcal{V}^1$ and $\mathcal{R}^1$ further by comparing the content of the transcripts and recipe steps. First, we tokenize the transcripts and recipe steps, followed by lemmatization, and select only content words. Then, we compute the token-IoU and token-recall for each video-recipe pair. Finally, we keep the pairs with token-IoU $\geq \lambda_{iou}$ and token-recall $\geq \lambda_{recall}$. We set $\lambda_{recall}=0.3$ and $\lambda_{iou}=0.1$ to obtain the refined video set $\mathcal{V}^2$, which yields $\approx$52K videos. Later, we split the videos into a train and validation set. Videos with $\lambda_{iou}\geq0.2$ form the validation set ($\approx3K$ videos), whereas the remaining $\approx$48K videos with $0.1 \leq \lambda_{iou} < 0.2$ form the training set. Similarly, we note the refined recipe set with $\mathcal{R}^2$. We use these thresholds to balance storage/computation resources and the resulting noise, however these hyper-parameters can be tuned to increase the size of the dataset. Figure~\ref{fig:data_create_stage2} (top) illustrates this part of the dataset creation.
\subsubsection{\sieveswap ASR Segments.} \label{sec:dataset_creation_3}
As mentioned earlier, ASR transcripts contain irrelevant content. Using a retrieval-based approach, we replace these transcripts with human-written instructional steps to improve the relevance of the text associated with the video segment. 
We retrieve the nearest recipe step $q \in \mathcal{S}, \ \forall \mathcal{S} \in \mathcal{R}^2$ for each individual ASR transcript segment $a$ (see Eq.~\ref{eq:asr_tuple}) based on the cosine similarity between their text embeddings (extracted with MPNet~\cite{song2020mpnet, reimers-2019-sentence-bert}). If the similarity is above a threshold, we swap $a$ with $q$. Formally, let $\kappa(a)$ be a function that returns the nearest recipe step $q$ for an ASR transcript $a$, where the similarity is calculated in the text embedding space. For a video $\mathbf{v} = (\mathcal{A}, T^v)$, recall that $\mathcal{A} = \{(a_j, t^s_j, t^e_j)\}$, we have (dropping $T^v$ for clarity): 
\begin{equation}
    \sieveswap({\mathbf{v}}) = \{(q_j, t^s_j, t^e_j)\} \big| \ q_j = \kappa(a_j) \ \land \ sim \big( \phi \langle a_j \rangle, \phi \langle q_j \rangle \big) \geq \lambda_{sim}
\end{equation}
where $\phi \langle . \rangle$ is the function to compute the embedding vector, $sim$ is the cosine similarity and $\lambda_{sim} = 0.75$. We discard a video segment if its corresponding ASR transcript has a neighbor whose similarity falls below this threshold. The collection of videos with human-written instructional steps is finally defined as:
\begin{equation}
    \mathcal{V}^{\sieveswap} = \{ \sieveswap(\mathbf{v}) | \ \forall \ \mathbf{v} \in \mathcal{V}^2 \}
\end{equation}
The size of $\mathcal{V}^{\sieveswap}$ remains the same as $\mathcal{V}^2$ in terms of videos; however, the number of video segments drops from 2.75M to 0.51M. With this procedure (depicted at the bottom of Figure~\ref{fig:data_create_stage2}), non-instructional parts of transcripts are removed since our text database contains only recipe steps. In contrast to prior methods, our selection of transcripts with their temporal boundaries helps to bridge the gap between the pre-training and downstream tasks and filters the irrelevant transcripts. Additional analysis of the \sieveswap dataset and qualitative examples are provided in the supplementary material.
\section{Procedure Transformer (ProcX)}
\label{sec:procx}
We now present our ProcX model which extendeds PDVC \cite{wang2021end}. We first discuss the core elements of PDVC and then introduce several architectural enhancements.

\subsubsection{Preliminary: Set-based Localization and Captioning.} The PDVC architecture has been inspired by the DETR model~\cite{carion2020end} and generates $\mathcal{N}$ proposals along with its captions using an LSTM to sequence events. Like DETR, it optimizes various losses using the Hungarian algorithm:
\begin{equation}
    \mathcal{L} = \beta_{box} \mathcal{L}_{box} + \beta_{cap} \mathcal{L}_{cap} + \beta_{fl} \mathcal{L}_{fl} + \beta_{count} \mathcal{L}_{count} 
\label{eq:pdvc_loss}
\end{equation}
where $\mathcal{L}_{cap}$ is the cross-entropy loss to predict the next token, $\mathcal{L}_{count}$ is the cross-entropy loss for segment count prediction, $\mathcal{L}_{fl}$ is the focal loss \cite{lin2017focal} between the positive and negative proposals, and $\mathcal{L}_{box}$ is the gIoU \cite{rezatofighi2019generalized} loss between the predicted and ground-truth temporal boundaries. More information about the base model and its architecture can be found in the supplementary material.

\subsubsection{Key-aware Deformable Attention.}
The PDVC employs vanilla deformable attention \cite{zhudeformable} to reduce the quadratic complexity of dense attention and achieve faster convergence. However, learning the attention weights based solely on the query by a linear projection (i.e., ignoring the query-key product) hampers the efficiency of cross-scale attention. As shown in Figure~\ref{fig:kda_analysis_attn}a, the entire scale collapses to focus on a single sampled point in the self-attention of the encoder module. Inspired by object detection techniques \cite{li2023lite}, we employ key-aware deformable attention for improved acquisition of multi-scale features, allowing diverse sampling across scales (see Figure~\ref{fig:kda_analysis_attn}b). More precisely, in vanilla deformable attention each head will sample $M$ points from multi-scale temporal features $S$ as value $V$ based on sampling offsets. The sampling offsets $\Delta p$ and their corresponding attention weights are directly predicted from queries using two linear projections denoted as $W_p$ and $W_a$. The sampled values $V$ and projected attention weights are utilized to compute the deformable attention:
\begin{figure}[t]
    \centering
    \includegraphics[width=0.9\linewidth]{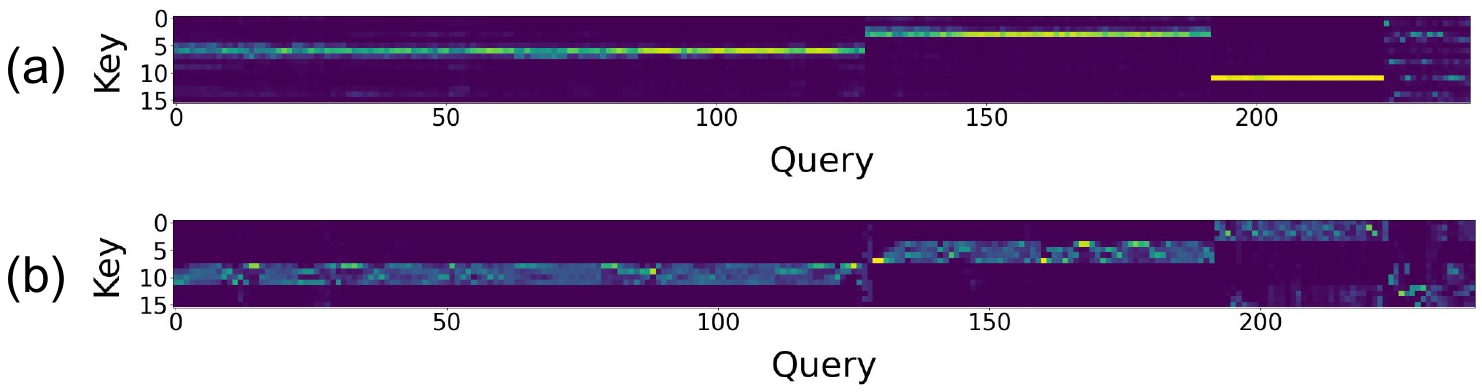}
    \captionsetup{font=scriptsize}
    \caption{Attention visualization of one head of the visual encoder. Attention is computed between multi-scale video features and sampled points across scales. In vanilla deformable attention \textbf{(a)}, the attention collapses to focus on a single sampled point and fails to attend to the entire video features. With our modified key-aware deformable attention \textbf{(b)}, the attention weights spread across the video and help to utilize the entire video context to learn better multi-scale video features.}
    \label{fig:kda_analysis_attn}
\end{figure}
\begin{align}
    \Delta p = Q\cdot W_p \quad \text{;} \quad  V = \Phi(S, p + \Delta p)\cdot W_v\\
    DeformAttn(Q, V) = \sigma(Q\cdot \highlight{W_a})\cdot V \label{eq:deform_attn}
\end{align}
where $\Phi$ is the sampling function with bi-linear interpolation and $\sigma$ is the softmax function. For the modified key-aware deformable attention, we propose to sample $M$ points for both keys and values from the multi-scale features and utilize standard scaled-dot product-based attention, as formulated below:
\begin{align}
    \highlight{K = \Phi(S, p + \Delta p)\cdot W_k} \quad \text{;} \quad  V = \Phi(S, p + \Delta p)\cdot W_v\\
    KeyDeformAttn(Q, V) = \sigma(Q\cdot \highlight{K^\intercal})\cdot V
    \label{eq:key_deform_attn}
\end{align}
where $W^k$ is a linear projection of sampled keys. The sampled key points participate in the key-aware deformable attention layer to weight the importance of each sampled value by comparing it with keys, as highlighted in Eq.~\ref{eq:deform_attn},~\ref{eq:key_deform_attn}.
\subsubsection{Contrastive Transformer.}
Due to the difficulty of optimizing the conditional likelihood objective for generating captions with vanilla transformer-based models, previous research employed denoising \cite{Luo2020UniVL, yang2023vid2seq} or contrastive losses \cite{yu2022coca}. We use the contrastive objective, which can be employed both in supervised settings and in pre-training, unlike the denoising objective, which is primarily utilized during pre-training. To learn better multi-modal video-to-text and text-to-video representations, we follow the decoupled architecture from the CoCa model \cite{yu2022coca} and employ the InfoNCE loss. However, it is difficult to establish a precise one-to-one correspondence between video and text during pre-training due to the misalignment between transcripts and video. We thus incorporate the MIL-InfoNCE loss \cite{miech2020end} during pre-training and use standard InfoNCE during finetuning.
\subsubsection{IoU-aware Confidence Score (gVFL).}
PDVC chooses the top-$k$ proposals based on predicted confidence scores during inference. Due to the presence of separate prediction heads, there is no relationship between the confidence score and the temporal boundary prediction in PDVC. This often results in higher confidence scores for proposals with poor temporal localization. We experimented with assuming oracle temporal localization during inference to confirm this observation. We employed the Hungarian Algorithm to establish a one-to-one mapping between predicted proposals and the ground-truth segments, yielding a substantial gain of $\approx$20\% in segmentation quality. Similar issues in object detection models have been addressed using an IoU-guided confidence loss function \cite{jiang2018acquisition, zhang2021varifocalnet}.
To address this issue in our setting, we replace the focal loss $\mathcal{L}_{fl}$ in Eq.~\ref{eq:pdvc_loss} with a modified varifocal loss \cite{zhang2021varifocalnet} to optimize the confidence scores. The standard varifocal loss employs the IoU between the predicted and ground-truth segments as a target for confidence score optimization. However, during the initial training stages, there might be no overlap between the proposal and ground truth segments, resulting in zero IoU. This can lead to sub-optimal confidence score optimization. To mitigate this, we use the generalized-IoU (gIoU) metric \cite{rezatofighi2019generalized} instead of standard IoU.  The gIoU calculates the distance between the predicted and ground-truth segments when there is no overlap (i.e., IoU = 0). More formally, our modified varifocal loss is:
\begin{align}
    \mathcal{L}_{g{\text -}vfl} = 
    \begin{cases}
        -g (g\log(p) + (1-g)\log(1-p)) & g > 0 \\
        -\alpha p^\gamma \log(1-g) & g = 0 
    \end{cases} 
\end{align}
where $p$ is the IoU-aware confidence score and $g$ is the scaled gIoU between the generated ($\hat{b}$) and ground-truth ($b$) segment, i.e., $g = 0.5 (1 + gIoU(b,\hat{b}))$.
\label{sec:method}

\section{Experimental Setup}
\subsection{Datasets}
We evaluate two instructional video datasets: \textbf{YouCook2} \cite{ZhXuCoCVPR18} and \textbf{Tasty} \cite{sener2019zero}. These datasets consist of cooking videos with detailed annotations specifying instruction steps in English sentences along with temporal boundaries. YouCook2 videos are characterized by an unconstrained environment and a third-person viewpoint, while Tasty videos are captured using an overhead camera. Another crucial difference between YouCook2 and Tasty is that Tasty does not include corresponding transcripts. This exemplifies the more realistic and challenging setting of not having step narrations. The assumption of having narrations at test time (as made in Vid2Seq~\cite{yang2023vid2seq}) is not only limiting but also turns the problem into a text-to-text translation rather than a video-to-language problem. Detailed information about the datasets is available in supplementary material.

\subsection{Evaluation Metrics}
We follow prior work \cite{wang2021end, batra2022closer, yang2023vid2seq} and employ the official evaluation tool from~\cite{krishna2017dense} to compute evaluation metrics. The tool first determines the matched pair between the ground truth and predicted segments based on maximum temporal overlap across IoU thresholds \{0.3, 0.5, 0.7, 0.9\}. Subsequently, it computes standard captioning metrics, i.e., METEOR \cite{banerjee-lavie-2005-meteor}, CIDER \cite{vedantam2015cider} and BLEU-4 \cite{papineni2002bleu} over the matched pairs. These metrics fail to account for the sequential coherence of steps, i.e., the extent to which the generated instructions cover the video's sequential progression. We thus additionally report SODA-C \cite{fujita2020soda} for instruction sequences and SODA-D \cite{batra2022closer} for segment localization. SODA-D takes into consideration the unique nature of non-overlapping segments within instructional videos. We prioritize SODA-C as the primary metric for its more suitable approach, employing temporally ordered and one-to-one matching between predicted and ground-truth instructions. In contrast, prior metrics use one-to-many matching at different IoU thresholds. SODA-C further evaluates text quality through METEOR and F-measure scores, while also penalizing redundancy in instructions/text.

\subsection{Implementation Details}
We follow \cite{wang2021end, batra2022closer} and pre-compute segment-level video features for both datasets at a fixed 16~fps, using S3D~\cite{miech2020end}. Features are resized to a fixed temporal size of 256 via nearest interpolation during pre-training and fine-tuning. For our ProcX model, we set the hyper-parameters for the visual encoder and the localization decoder following PDVC~\cite{wang2021end}. However, we enhance the capacity of the model by incorporating four layers in both the visual encoder and the localization decoder. To train the instruction generation model we learn a customized vocabulary comprising 10K tokens using Huggingface \cite{wolf2019huggingface} and employing the WordPiece tokenization algorithm \cite{song2020fast}. Further details in the supplementary material.
\section{Experimental Evaluation}
\subsection{Ablation Studies}
\subsubsection{\sieveswapnospace.} 
We developed \sieveswap to automatically curate a smaller dataset for pre-training models with a limited budget. Here we validate our approach and report results obtained by training the model on raw transcripts. In Section~\ref{sec:dataset_creation} this is the dataset compiled after sieving videos with title and content overlap of words between recipes and ASR text, denoted with $\mathcal{D}^v_2$. It is important to note that videos here still contain raw ASR text segments. Table~\ref{table:dvc_ablate_pretrain} compares results obtained pre-training on this dataset (Raw ASR) to the dataset obtained after filtering ASR text segments and refining them with \sieveswapnospace. All models pre-trained with \sieveswap transcripts improve consistently across all metrics despite being pre-trained only on a fifth of the data. This demonstrates the effectiveness of our approach: we improve performance by curating a smaller and better pre-training dataset. We reduce noise by discarding irrelevant ASR transcripts (sieve stage), often filled with sentences unrelated to the instruction (e.g., greetings) and replacing instructional ASR transcripts with recipe steps (swap stage). We show additional ablations on using only-sieved ASR along with combining sieved ASR text and \sieveswap text in the supplementary material.
\begin{table}[tb]
\scriptsize
\renewcommand{\tabcolsep}{0.1cm}
\captionsetup{font=scriptsize}
\caption{Effectiveness of \sieveswapnospace. Models were finetuned on YouCook2 and pre-trained with raw ASR transcripts or \sieveswap Captions. Segments are the number of video segments (and corresponding captions) used for pre-training. With \sieveswap the pre-training data points are reduced by a factor of five, but models still achieve better performance thanks to reduced noise and better captions. We highlight the primary metric SODA-C. Best results per section are in bold, while best results across all models are underlined.}
    \label{table:dvc_ablate_pretrain}
    \centering
    \resizebox{\textwidth}{!}{%
    \begin{tabular}{lcccacccc}
        \toprule[1.5pt]
        \specialcell{\bf Method} & \specialcell{\bf Visual} & \multicolumn{1}{c}{\specialcell{\bf Segments}} & \specialcell{\bf Captions} & \multicolumn{5}{c}{\specialcell{\bf YouCook2}}  \\
        \cmidrule(lr){5-9}
            & {\bf Features} & (in Million) & & SODA-C & METEOR & CIDER & B4 & SODA-D \\ 
            \midrule
            Vid2Seq \cite{yang2023vid2seq} & CLIP & 2.75 & Raw ASR & 3.65 & 3.89 & 14.57 & 0.52 & 29.64 \\
            Vid2Seq & CLIP & \textbf{0.51} & Sieve-\&-Swap (Ours) & \textbf{4.21} & \textbf{4.19} & \textbf{15.86} & \textbf{0.61} & \textbf{29.71} \\ \midrule
            PDVC \cite{wang2021end} & S3D & 2.75 & Raw ASR & 5.81 & 6.27 & 32.44 & 1.26 & 32.71  \\
            PDVC & S3D & \textbf{0.51} & Sieve-\&-Swap (Ours) & \textbf{6.16} & \underline{\textbf{6.85}} & \textbf{36.27} & \textbf{1.61} & \textbf{32.80} \\ \midrule
            ProcX (Ours) & S3D & 2.75 & Raw ASR & 6.97 & 6.34 & 38.23 & 1.71 & 37.08 \\
            ProcX (Ours) & S3D & \textbf{0.51} & Sieve-\&-Swap (Ours) & \underline{\textbf{7.29}} & \textbf{6.76} & \underline{\textbf{40.57}} & \underline{\textbf{1.92}} & \underline{\textbf{37.61}} \\
    \bottomrule[1.5pt]
    \end{tabular}
    }
\end{table}
\begin{table}[!htb]
    \captionsetup{font=scriptsize}
    \caption{\textbf{(a}) Shows the effect of the adding Transformer and modified gIoU guided varifocal loss (gVFL) to PDVC~\cite{wang2021end}. \textbf{(b)} Shows the effect of the Key-Aware Deformable Attention in the video encoder and segmentation decoder. }
    \label{table:dvc_ablate_procx}
    \begin{adjustbox}{width=1\linewidth}
    \begin{subtable}{1\linewidth}
        \renewcommand{\tabcolsep}{0.15cm}
        \centering
        \captionsetup{font=normalsize}
        \caption{}
        \label{table:dvc_ablate_losses}
        \begin{tabular}{lcacccc}
            \toprule[1.5pt]
            \specialcell{\bf Captioner} & \specialcell{\bf gVFL} & \multicolumn{5}{c}{\specialcell{\bf YouCook2}} \\
            \cmidrule(lr){3-7}
            & & SODA-C & METEOR & CIDER & B4 & SODA-D \\ \midrule
            LSTM &  & 5.43 & 5.40 & 27.42 & 0.85 & 31.48 \\
            Transformer & \xmark & 4.94 & \textbf{5.67} & 28.88 & 1.19 & 28.64 \\
            Transformer & \cmark & \textbf{5.84} & 5.34 & \textbf{30.83} & \textbf{1.28} & \textbf{33.33} \\
        \bottomrule[1.5pt]
        \end{tabular}
    \end{subtable}%
    \begin{subtable}{1\linewidth}
        \renewcommand{\tabcolsep}{0.15cm}
        \centering
        \captionsetup{font=normalsize}
        \caption{}
        \label{table:dvc_ablate_kda}
        \begin{tabular}{ccacccc}
            \toprule[1.5pt]
            \multicolumn{2}{c}{\specialcell{\bf Key Aware}} & \multicolumn{5}{c}{\specialcell{\bf YouCook2}} \\
            \cmidrule(lr){1-2}\cmidrule(lr){3-7}        
            Encoder & Decoder & SODA-C & METEOR & CIDER & B4 & SODA-D \\ \midrule
            \xmark & \xmark & 5.84 & 5.34 & 30.83 & 1.28 & 33.33 \\
            \cmark & \xmark & 5.95 & \textbf{5.40} & \textbf{31.20} & \textbf{1.29} & 33.26 \\
            \cmark & \cmark & \textbf{6.03} & 5.35 & 30.84 & 1.27 & \textbf{34.63} \\
        \bottomrule[1.5pt]
        \end{tabular}
    \end{subtable} 
    \end{adjustbox}
\end{table}

\subsubsection{ProcX Architecture Improvements.}
We systematically study the integration of the contrastive transformer and the gIoU-guided varifocal loss (g-VFL) to PDVC~\cite{wang2021end}. In Table~\ref{table:dvc_ablate_losses} we show the performance of PDVC~\cite{wang2021end} with the LSTM captioner as well as the Transformer captioner with and without gVFL. Replacing the LSTM with a contrastive transformer improves caption generation but reduces SODA-D, leading to lower SODA-C scores. We attribute this performance drop to the utilization of a low confidence score for selecting top-$k$ segments. To address this, we introduce a modified varifocal loss with generalized-IoU guidance to enhance the confidence score, improving our ProcX model's performance. Table~\ref{table:dvc_ablate_kda} demonstrates the effectiveness of enhanced key-aware deformable attention in the visual encoder and segment localization decoder. By integrating the key-aware attention mechanism, we observe notable performance improvements in both self-attention during video encoding and cross-attention in the localization decoder, reflected in SODA-D. There is a marginal change in the other \textit{caption} metrics as we introduce this attention mechanism in the \textit{visual} encoding and decoding. The marginal gains from individual components are due to the small size of YouCook2 (0.8K), but overall we observe an improvement in SODA-C by 11.05\% and 50.83\% on YouCook2 and Tasty (3K) datasets (rows 3, 4 in Table~\ref{table:dvcsota_ft}).
\begin{table}[tb]
\captionsetup{font=scriptsize}
\caption{Comparison of our results on YouCook2 and Tasty with state-of-the-art models. All models receive visual input, but those in gray rows get both visual and ASR text input. Vid2Seq$^{\ddagger}$ and PDVC$^{\ddagger}$: results reported from~\cite{yang2023vidchapters}. Vid2Seq$^{\dagger}$ and PDVC$^{\dagger}$: Results from our run. The evaluation metrics are an average of three different runs for our model.}
    \label{table:dvcsota_ft}
    \renewcommand{\tabcolsep}{0.1cm}
    \centering
    \resizebox{.99\textwidth}{!}{%
    \begin{tabular}{clccrcaccccacccc}
        \toprule[1.5pt]
        \specialcell{\bf } & \specialcell{\bf Method}  & \specialcell{\bf Video} & \specialcell{\bf ASR} & \multicolumn{2}{c}{\specialcell{\bf Pre-training}} & \multicolumn{5}{c}{\specialcell{\bf YouCook2 (val)}} & \multicolumn{5}{c}{\specialcell{\bf Tasty (test)}}  \\
        \cmidrule(lr){5-6}\cmidrule(lr){7-11}\cmidrule(lr){12-16}
         & & {\bf Features} & {\bf Text} & Seg (M) & Captions & SODA-C & METEOR & CIDER & B4 & SODA-D & SODA-C & METEOR & CIDER & B4 & SODA-D \\ 
         \midrule
            \multicolumn{16}{l}{\specialcell{\textit{\color{blue}{Without PreTraining}}}} \\
            1. & Vid2Seq$^{\dagger}$ \cite{yang2023vid2seq} & CLIP & \xmark & - & - & 3.61 & 3.83 & 14.24 & 0.42 & 27.16 & 2.58 & 3.03 & 9.48 & 0.38 & 35.88 \\
            2. & PDVC \cite{wang2021end} & CLIP & \xmark & - & - & 4.35 & 4.27 & 19.44 & 0.44 & 27.29 & 7.82 & 8.39 & 40.03 & 3.09 & 43.61 \\
            3. & PDVC \cite{wang2021end} & S3D & \xmark & - & - & 5.43 & \textbf{5.40} & 27.42 & 0.85 & 31.63 & 7.18 & 7.87 & 37.32 & 2.61 & 43.51 \\
            4. & ProcX (ours) & S3D & \xmark & - & - & \textbf{6.03} & 5.35 & \textbf{30.84} & \textbf{1.27} & \textbf{34.63} & \textbf{10.83} & \textbf{10.55} & \textbf{70.90} & \textbf{6.26} & \textbf{51.36} \\
            \midrule 
            \multicolumn{16}{l}{\specialcell{\textit{\color{blue}{With PreTraining}}}} \\
            5. & Vid2Seq$^{\ddagger}$ \cite{yang2023vid2seq} & CLIP & \xmark & $\sim$900.0 & Raw ASR & 5.70 & - & 25.30 & - & - & - & - & - & - & - \\
            6. & Vid2Seq$^{\dagger}$ & CLIP & \xmark & 0.5 & \sieveswap (ours) & 4.21 & 4.19 & 15.86 & 0.61 & 29.71 & 7.28 & 8.57 & 46.29 & 4.78 & 38.94 \\
            7. & PDVC$^{\ddagger}$ \cite{wang2021end} & CLIP & \xmark & 7.0 & Video Chapters & 5.90 & \textbf{7.50} & 34.70 & - & - & - & - & - & - & - \\
            8. & PDVC$^{\dagger}$ & CLIP & \xmark & 0.5 & \sieveswap (ours) & 5.48 & 5.83 & 30.48 & 1.33 & 29.78 & 8.01 & 8.72 & 43.68 & 3.51 & 44.71 \\
            9. & PDVC$^{\dagger}$ & S3D & \xmark & 0.5 & \sieveswap (ours) & 6.16 & 6.85 & 36.27 & 1.61 & 32.80 & 7.64 & 8.31 & 40.97 & 2.84 & 44.94 \\
            10. & ProcX (ours) & S3D & \xmark & 0.5 & \sieveswap (ours) & \textbf{7.29} & 6.76 & \textbf{40.57} & \textbf{1.92} & \textbf{37.61} & \textbf{11.66} & \textbf{11.24} & \textbf{78.13} & \textbf{7.09} & \textbf{52.16} \\
            \midrule
            \rowcolor{gray!8}\gray{11.} & \gray{Vid2Seq$^{\ddagger}$} \cite{yang2023vid2seq} & \gray{CLIP} & \gray{\cmark} & \gray{136.0} & \gray{Raw ASR} & \gray{8.60} & \gray{10.50} & \gray{53.20} & \gray{-} & \gray{-} & \gray{-} & \gray{-} & \gray{-} & \gray{-} & \gray{-} \\
            \rowcolor{gray!8} \gray{12.} & \gray{Vid2Seq$^{\ddagger}$} \cite{yang2023vid2seq} & \gray{CLIP} & \gray{\cmark} & \gray{$\sim$143.0} & \gray{Raw ASR + VC} &  \gray{10.30} & \gray{12.30} & \gray{67.20} & \gray{-} & \gray{-} & \gray{-} & \gray{-} & \gray{-} & \gray{-} & \gray{-} \\
    \bottomrule[1.5pt]
    \end{tabular}
    }
\end{table}
\subsection{Comparison to State of the Art}
In the top section of Table~\ref{table:dvcsota_ft} (rows 1--4), we compare results without pre-training on two datasets, i.e., models are directly fine-tuned on YouCook2 and Tasty. Set-based transformers such as PDVC \cite{batra2022closer} and our ProcX demonstrate superior performance compared to the sequence-to-sequence model Vid2Seq\cite{yang2023vid2seq}. ProcX consistently outperforms PDVC, underscoring the effectiveness of an end-to-end transformer-based approach in comparison to an LSTM-based model. ProcX improves SODA-C by 0.6 and 3 points on YouCook2 and Tasty, indicating higher temporal coherence in the generated instructions. ProcX also enhances localization performance (SODA-D) by 3 and 8 points on the two datasets. 

In the middle section of Table~\ref{table:dvcsota_ft} (rows 5--10), we present results obtained from pre-training models on various datasets and fine-tuning them on the respective downstream datasets. ProcX pre-trained on our \sieveswap dataset (row 10) achieves a new state-of-the-art, despite being pre-trained on only $0.5$ million segments, as opposed to $\approx$900 million (row 5) or $7$ million (row 7). This shows again both the effectiveness of our \sieveswap method to compile a better pre-training dataset and the efficiency of ProcX. Lastly, the bottom section of Table~\ref{table:dvcsota_ft} (rows 11--12) includes results from Vid2Seq obtained using transcripts during both training and inference. We note a large gap with all video-only methods, however results are not directly comparable since using ASR text as input during inference gives a substantial advantage for this task. This is because ASR text has a significant content overlap with the generated captions and additionally provides a temporal segmentation of the video.
\subsubsection{Zero Shot Procedure Learning.} 
We also test models in a zero-shot setting, where they are only pre-trained and not fine-tuned on the downstream datasets. Here we pre-train models on our \sieveswap dataset as well as Video Chapters~\cite{yang2023vidchapters} (for Vid2Seq) and a subset of cooking videos from HowTo100M with raw ASR obtained after sieving (see Section~\ref{sec:dataset_creation_2}). This experiment is reported in Table~\ref{table:dvcsota_zs}. ProcX consistently enhances caption metrics across both datasets, highlighting the efficacy of our new model. Importantly, all models pre-trained on \sieveswap dataset perform better than they do when pre-trained on the much larger Video Chapters \cite{yang2023vidchapters} (7M~segments) and the noisier raw ASR subset, demonstrating the effective generalization of the \sieveswap dataset for different models. We ascribe this to the fact that the \sieveswap dataset helps in reducing the domain gap between pre-training text (transcribed speech) and target text (human written instructions). We created the \sieveswap dataset with the goal of enhancing caption quality. However, we noticed that it also contributes to improved localization on YouCook2. The performance drop on the Tasty dataset for localization can be attributed to a mismatch in segment duration. Specifically, the average segment duration in Tasty is six seconds, closely matching with shorter transcript segments.
\begin{table}[tb]
\captionsetup{font=scriptsize}
\renewcommand{\tabcolsep}{0.1cm}
\caption{Zero-shot performance on YouCook2 and Tasty. We pre-train models without fine-tuning them on the target datasets. Vid2Seq$^{\ddagger}$ refers to the official results from VidChapters \cite{yang2023vidchapters}. The results per section are in bold, while best result across all models are highlighted with underline.}
    \label{table:dvcsota_zs}
    \centering
    \resizebox{0.99\textwidth}{!}{%
    \begin{tabular}{lcccaccccacccc}
        \toprule[1.5pt]
        \specialcell{\bf Method} & \specialcell{\bf Video} & \multicolumn{2}{c}{\specialcell{\bf Pre-training}} & \multicolumn{5}{c}{\specialcell{\bf YouCook2 (val)}} & \multicolumn{5}{c}{\specialcell{\bf Tasty (test)}} \\
        \cmidrule(lr){3-4}\cmidrule(lr){5-9}\cmidrule(lr){10-14}
             & {\bf Features} & \# Segments & Captions & SODA-C & METEOR & CIDER & B4 & SODA-D & SODA-C & METEOR & CIDER & B4 & SODA-D \\ \midrule
             
            Vid2Seq$^{\ddagger}$ \cite{yang2023vid2seq} & CLIP & 7.0 & Video Chapters \cite{yang2023vidchapters} & 0.70 & 0.50 & 1.10 & - & - & - & - & - & - & - \\
            Vid2Seq$^{\dagger}$ \cite{yang2023vid2seq} & CLIP & 2.5 & Raw ASR & 0.25 & 0.35 & 0.64 & 0.01 & 4.63 & 0.91 & 1.28 & 1.78 & 0.01 & 25.12 \\
            Vid2Seq$^{\dagger}$ & CLIP & 0.5 & \sieveswap (ours) & \textbf{1.33} & \textbf{1.73} & \textbf{3.83} & \textbf{0.04} & \textbf{17.52} & \underline{\textbf{3.32}} & \textbf{4.52} & \textbf{12.52} & \textbf{0.56} & \underline{\textbf{37.55}} \\ \midrule
            PDVC \cite{wang2021end} & S3D & 2.5 & Raw ASR & 0.72 & 0.78 & 2.39 & 0.00 & 10.51 & 2.27 & 2.78 & 6.82 & 0.13 & \textbf{31.32} \\
            PDVC  & S3D & 0.5 & \sieveswap (ours) & \textbf{2.01} & \textbf{2.99} & \underline{\textbf{9.78}} & \textbf{0.17} & \textbf{21.70} & \textbf{2.64} & \textbf{4.18} & \textbf{12.94} & \textbf{0.58} & 25.52 \\ \midrule
            ProcX (ours) & S3D & 2.5 & Raw ASR & 1.10 & 1.14 & 2.86 & 0.02 & 17.83 & 2.65 & 3.41 & 7.93 & 0.35 & \textbf{35.34}  \\
            ProcX (ours) & S3D & 0.5 & \sieveswap (ours) & \underline{\textbf{2.03}} & \underline{\textbf{3.15}} & \textbf{7.75} & \underline{\textbf{0.27}} & \underline{\textbf{22.37}} & \textbf{2.82} & \underline{\textbf{6.24}} & \underline{\textbf{19.19}} & \underline{\textbf{1.36}} & 24.97 \\
    \bottomrule[1.5pt]
    \end{tabular}
    }
\end{table}

\section{Limitations and Future Work}
\sieveswap achieves state-of-the-art results with limited data and promises further gains by scaling the curated dataset and pre-training with larger compute. However, resource constraints currently hinder scaling, deferring it to future endeavors. Additionally, \sieveswap is generic and can be extended to other domains beyond cooking, e.g., leveraging WikiHow or utilizing domain transfer learning~\cite{li2022gain} from cooking instructional videos to general instructional videos.

\section{Conclusion}
\label{sec:conclusion}
Understanding procedural videos is a fundamental problem with a range of growing applications. It is challenging and requires solving two tasks: learning to localize steps and generating coherent descriptions for these steps. Models that solve this task need to have a high capacity and therefore require large amounts of training data. Existing pre-training approaches leverage video transcripts to create very large noisy datasets, which take a lot of time to train and have a domain gap between transcribed speech and written step description. To address these issues, we proposed \sieveswap, a technique to curate instructional video datasets leveraging language-only datasets of the same domain. The resulting dataset allows us to develop and train a novel large model, ProcX, using three orders of magnitude less data and surpassing existing state-of-the-art performance. Our curation approach is general and will open up new possibilities for training large video understanding models.

\myparagraph{Acknowledgements.} This work was supported in part by the UKRI Centre for Doctoral Training in Natural Language Processing, funded by UKRI grant EP/S022481/1 and the University of Edinburgh, School of Informatics. MR was funded in part by an Alexander von Humboldt Professorship in Multimodal Reliable AI sponsored by Germany's Federal Ministry for Education and Research. 

%
%
\bibliographystyle{splncs04}
\bibliography{main}

\clearpage
\appendix
\section*{Appendix}
The supplementary material contains the following:
\begin{enumerate}
    \item Additional Details of the \sieveswap dataset (Section~\ref{sec:sup_data});
    \item Additional Ablations (Section~\ref{sec:sup_ablations});
    \item Qualitative Results (Section~\ref{sec:sup_results});
    \item Architecture Details (Section~\ref{sec:sup_procx});
    \item Additional Evaluation Dataset and Implementation details (Section~\ref{sec:sup_impl}).
\end{enumerate}
\section{\sieveswap Dataset}\label{sec:sup_data}
\subsection{Analyzing Text Content} 
\begin{figure}[h]
    \centering
    \includegraphics[width=0.8\linewidth]{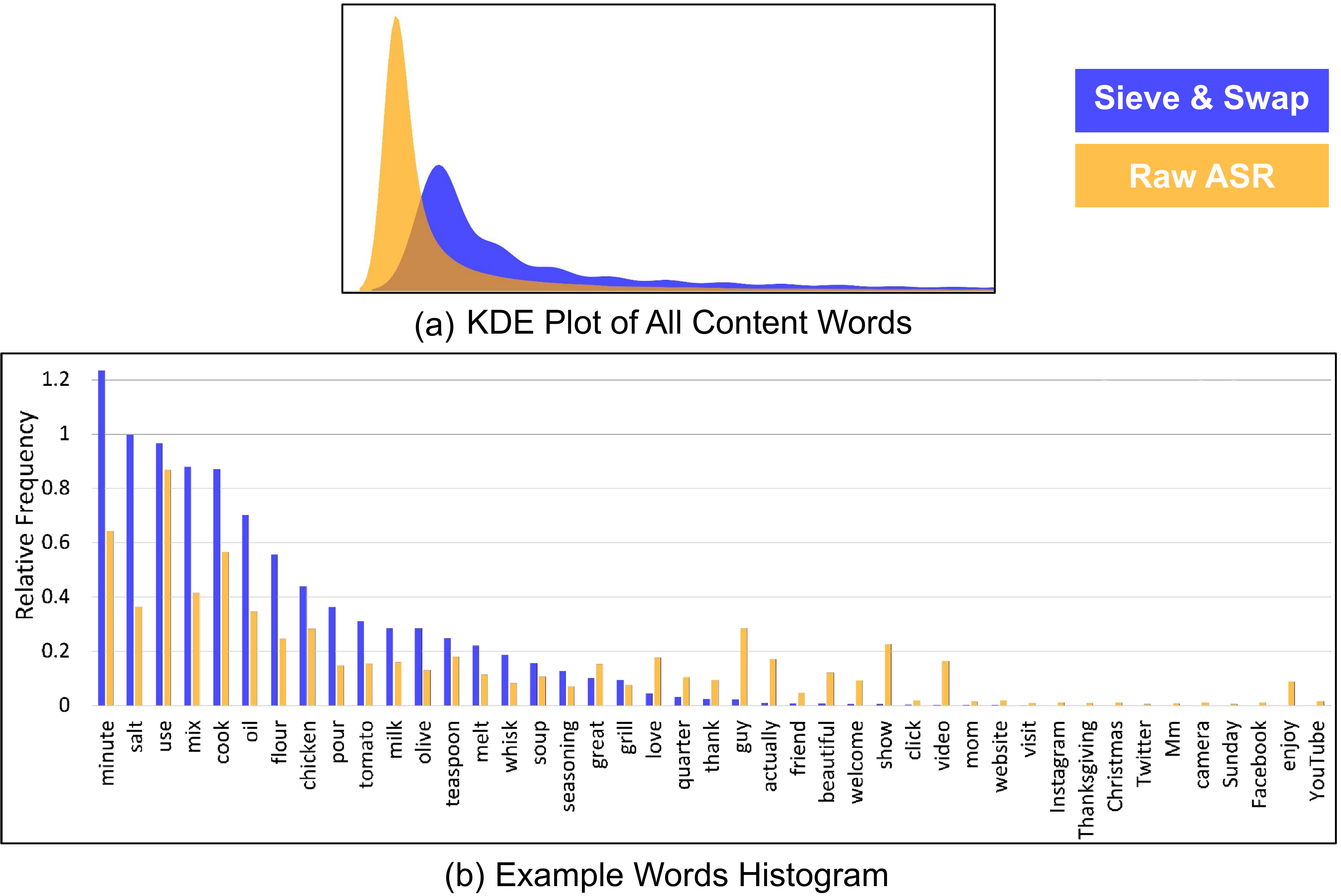}
    \captionsetup{font=scriptsize}
    \caption{(a) Kernel Density Estimation for Content words in raw ASR transcripts and our \sieveswap dataset. (b) Frequency histogram for a sample of words.}
\label{fig:sns_htc_dist}
\end{figure}
Here we analyze the text content before and after filtering ASR transcripts. We tokenize the raw ASR and \sieveswap text, i.e., we
perform lemmatization and select only content words (nouns, verbs, adverbs, adjectives, numerals, units) and compare their distributions. Figure~\ref{fig:sns_htc_dist} (a) shows the distribution of content words in the two datasets, where we can see that text with \sieveswap is more diverse as it contains more content words (the KDE is more spread for \sieveswap than it is for the Raw ASR). ASR transcripts include generic words (e.g., `subscribe') and utterance fillers (e.g., `Mm'). After filtering with \sieveswap these should decrease, while instructional words should increase. To show this, we compare a few words from the raw transcripts dataset and the refined \sieveswap dataset. In Figure~\ref{fig:sns_htc_dist} (b) we see that the count of instructional content words such as `add', `salt', `oil' increases, while the count for irrelevant words such as `friend', `Mm', `subscribe' decreases or drops to zero. These graphs show that indeed raw ASR transcript contain irrelevant content and that our \sieveswap procedure effectively reduces it.
\begin{figure}[h]
    \centering
    \includegraphics[width=0.95\linewidth]{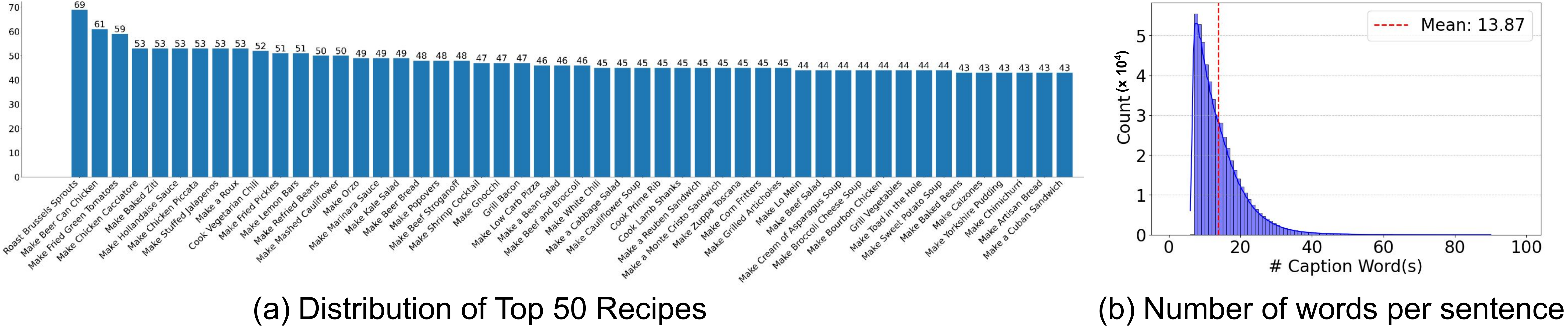}
    \captionsetup{font=scriptsize}
    \vspace{-0.1cm}
    \caption{(a) Top 50 recipes from the \sieveswap dataset containing 4K unique recipes. (b) Distribution of the number of words per instruction sentence.}
\label{fig:sns_htc_stats}
\end{figure}
\subsection{Additional Details}
\begin{figure}[h]
    \centering
    \captionsetup{font=scriptsize}
    \includegraphics[width=0.99\linewidth]{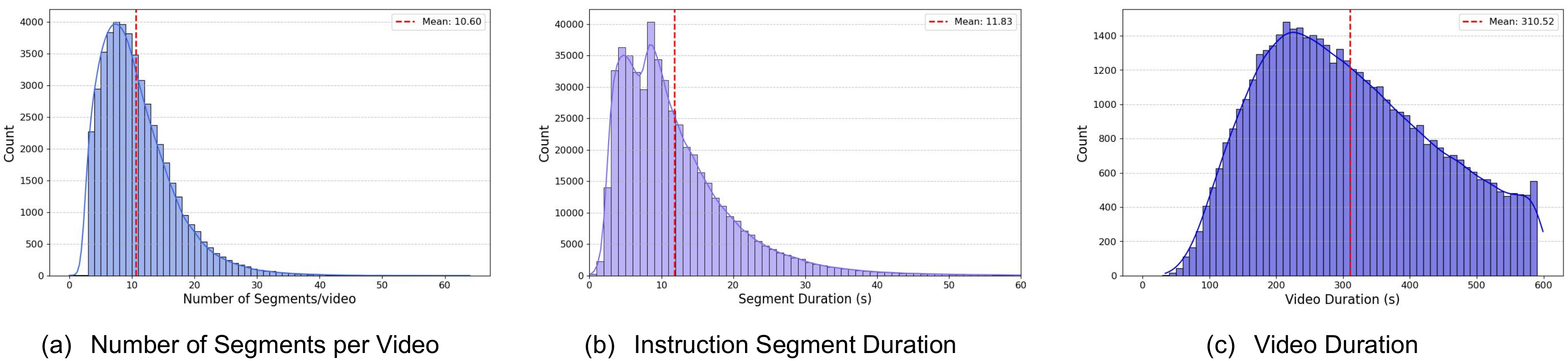}
    \caption{Statistics of \sieveswap. (a) Distribution of the number of steps per video (average: 10.6 steps), (b) Distribution of instruction segment duration (average: 11.83 seconds), and (c) Distribution of the video duration (average: 310.5 seconds).}
    \label{fig:htc_stats}
\end{figure}

\begin{figure*}[t]
    \centering
    \captionsetup{font=scriptsize}
    \includegraphics[width=0.99\linewidth]{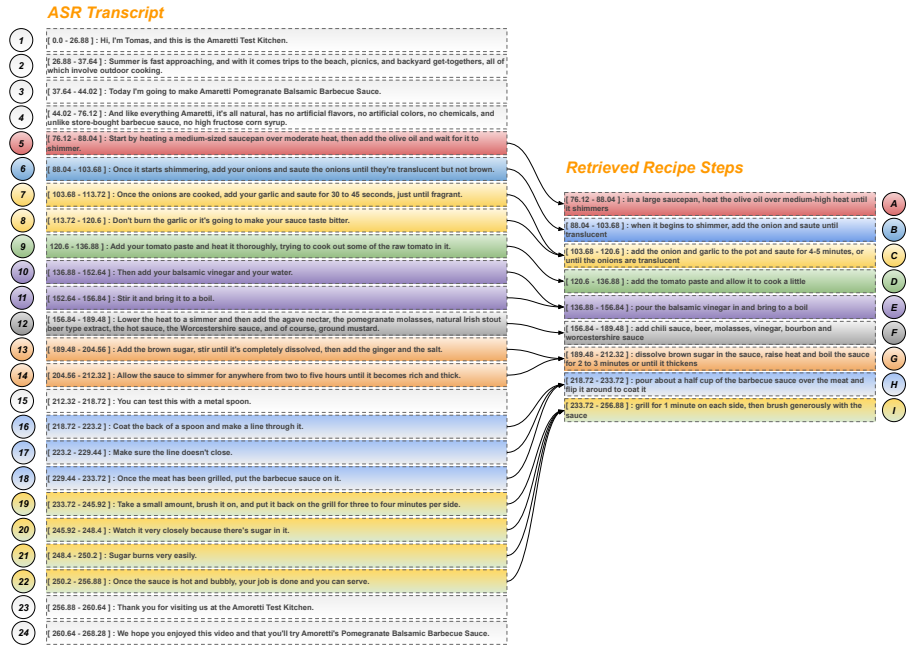}
    \caption{An instance extracted from our \sieveswap dataset, depicting the transcripts of a video along with the retrieved recipe steps following our  approach involving transcript filtering and replacement with closely matching steps. Matched transcripts - recipe step pairs are color coded. For example, transcript labeled with `5' is replaced with recipe step `A' and transcripts labeled with `7', `8' are replaced with recipe step `C'. The un-matched transcripts are filtered out from the final sample. The Video is sampled from \cite[HowTo100M]{miech2019howto100m} with YouTube ID:$tTithhQYkxE$. Best visualized in color.}\label{fig:htc_sample1}
\end{figure*}

\begin{figure*}[t]
    \centering
    \captionsetup{font=scriptsize}
    \includegraphics[width=0.99\linewidth]{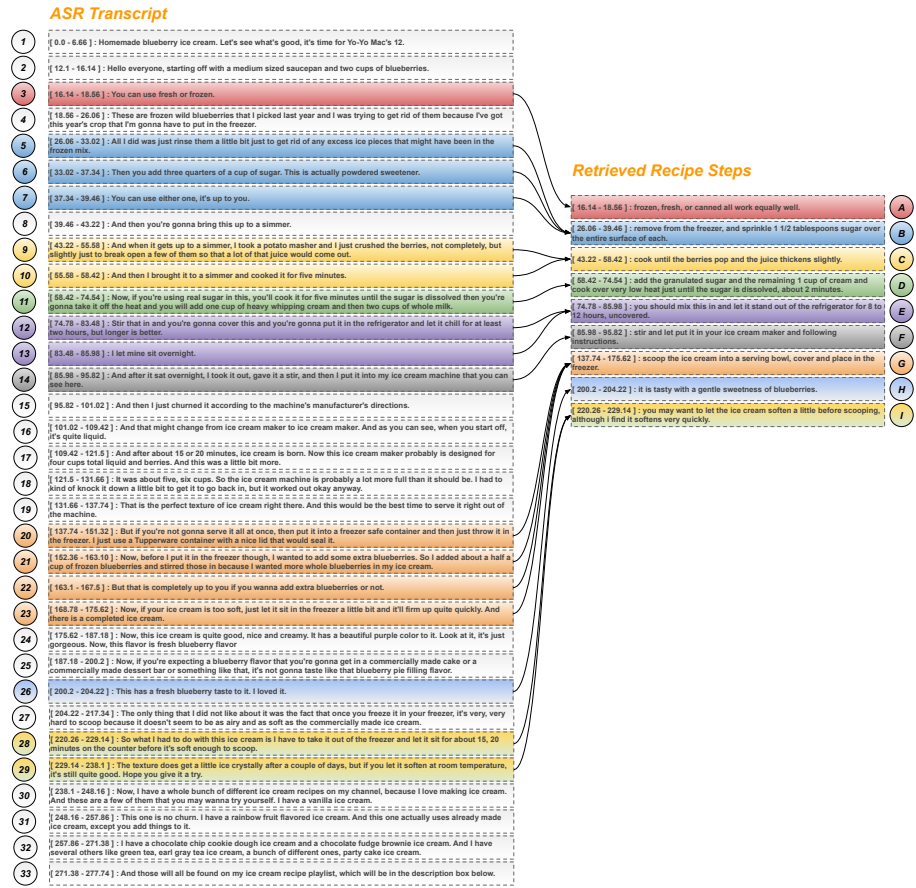}
    \caption{\textbf{\textit{Additional Example - 2}} from \sieveswap dataset with YouTube Video ID:$pehsD32ccDY$. Best visualized in color.}\label{fig:htc_sample2}
\end{figure*}

\begin{figure*}[t]
    \centering
    \captionsetup{font=scriptsize}
    \includegraphics[width=0.99\linewidth]{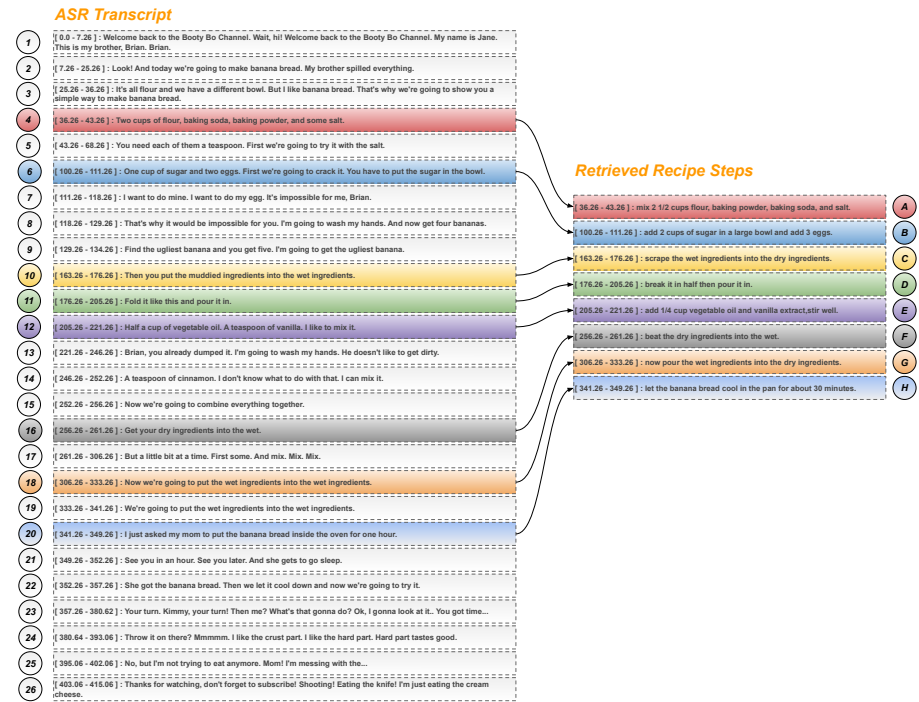}
    \caption{\textbf{\textit{Additional Example - 3}} from \sieveswap dataset with YouTube Video ID:$pehsD32ccDY$. Best visualized in color.}\label{fig:htc_sample3}
\end{figure*}

\begin{figure*}[t]
    \centering
    \captionsetup{font=scriptsize}
    \includegraphics[width=0.99\linewidth]{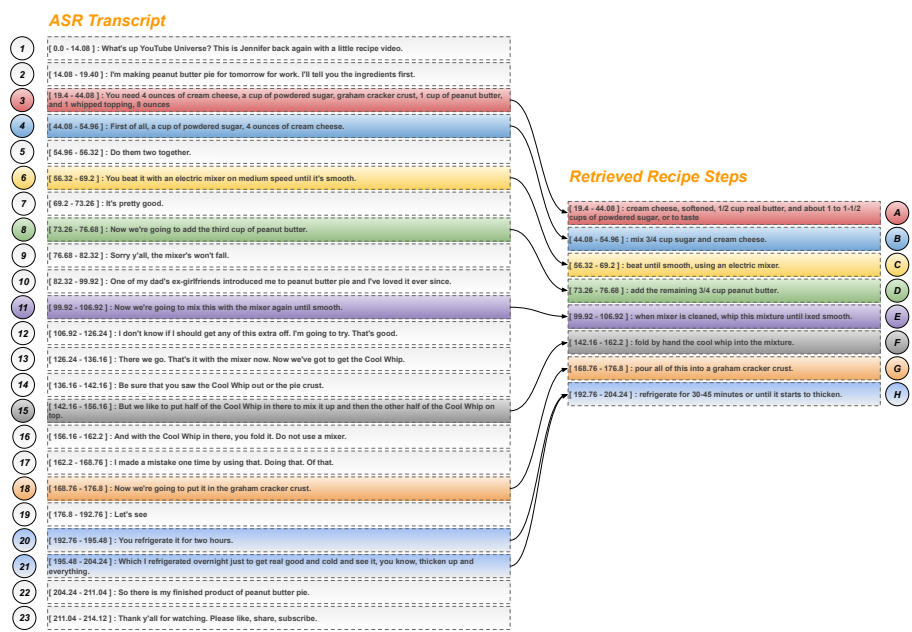}
    \caption{\textbf{\textit{Additional Example - 4}} from \sieveswap dataset with YouTube Video ID:$NqtS1q59oWg$. Best visualized in color.}\label{fig:htc_sample4}
\end{figure*}

\sieveswap comprises diverse procedures, totaling 4,109 distinct instructional recipes. In Figure~\ref{fig:sns_htc_stats} we illustrate (a) the histogram count of the top 50 recipes in the dataset (there are multiple recipes per title) and (b) the histogram count of the number of words per sentence. Additional statistics are depicted in Figure~\ref{fig:htc_stats}.  Figures \ref{fig:htc_sample1},~\ref{fig:htc_sample2}, \ref{fig:htc_sample3}, \ref{fig:htc_sample4} illustrate a few examples from our dataset. The left-hand side shows raw ASR transcripts along with their start/end times, while the right-hand side shows the retrieved recipe instructions substituting the respective transcript. Irrelevant transcripts (in gray) are general discussions that pertain to noisy segments, which are removed in \sieveswap. Note that we merge contiguous transcripts and replace them with a single recipe step. Specifically, we merge two consecutive segments shorter than 8 seconds only when the time gap between them is less than 4 seconds.

\begin{figure}[h]
    \centering
    \scriptsize
    \subfloat[\centering]{{\includegraphics[width=6cm]{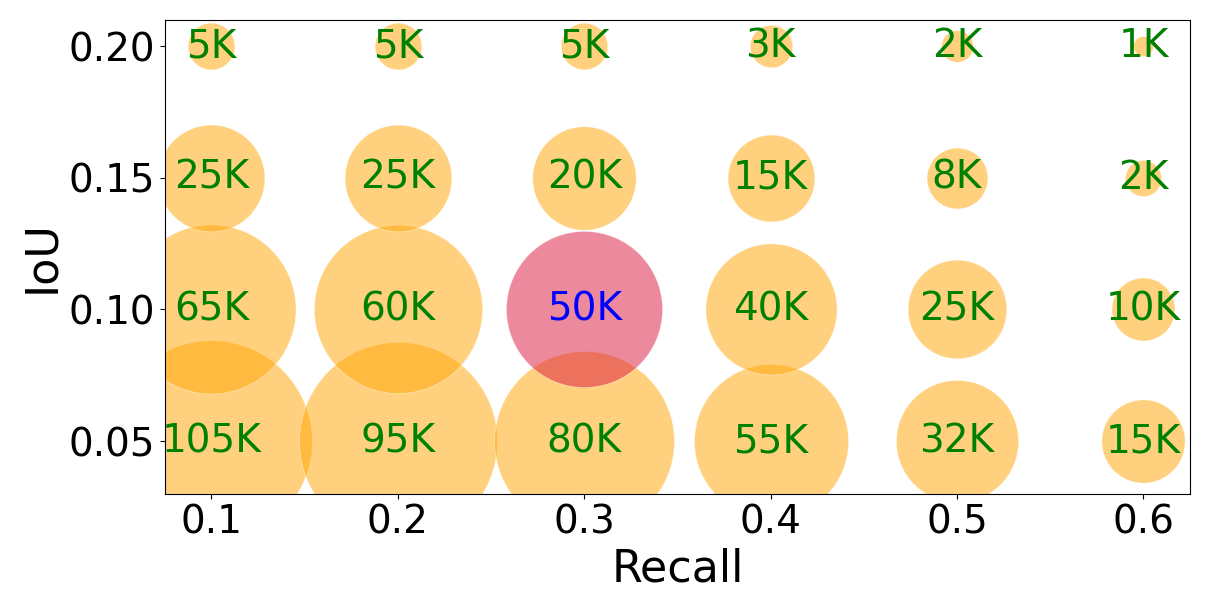} }}%
    \qquad
    \subfloat[\centering]{{\includegraphics[width=4.5cm]{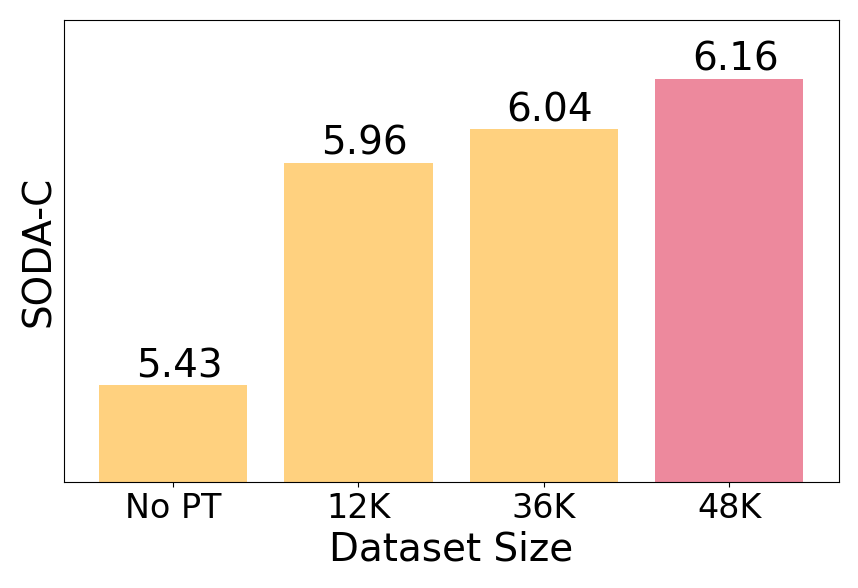} }}%
    \captionsetup{font=scriptsize}
    \vspace{-1.2em}
    \caption{\textbf{(a)} Size of the dataset w.r.t. $\lambda_{IoU}$ and $\lambda_{recall}$, \textbf{(b)} Ablation results with PDVC on YouCook2 by varying \sieveswap dataset size. \textbf{No PT}: No Pre-Training.}%
    \label{fig:size-lambda}
\end{figure}
\begin{table}[h]
    \captionsetup{font=scriptsize}
    \caption{SODA-C results on YouCook2. (a) \textbf{RA}: Raw ASR, \textbf{OS}: Only Sieve, \textbf{SS}: \sieveswap (ours). (b) \textbf{Mix}: Mix of SS \& OS (ours). (c) \textbf{VC}: Video Chapters~\cite{yang2023vidchapters}.}
    \label{table:rebuttal_ablate_procx}
    \begin{subtable}{0.3\linewidth}
        \scriptsize
        \renewcommand{\tabcolsep}{0.08cm}
        \renewcommand{\arraystretch}{0.6}
        \centering
        \begin{tabular}{ccc}
            \toprule[1pt]
            \multicolumn{3}{c}{\specialcell{\bf ProcX (S3D)}}\\
            \cmidrule(lr){1-3}
                RA & OS & SS(our) \\ 
                \midrule
                6.97 & 6.93 & 7.29 \\
        \bottomrule[1pt]
        \end{tabular}
        \captionsetup{font=scriptsize}
        \caption{}
        \label{table:rebutal_ablation_a}
    \end{subtable}%
    \begin{subtable}{0.43\linewidth}
        \scriptsize
        \renewcommand{\tabcolsep}{0.08cm}
        \renewcommand{\arraystretch}{0.6}
        \centering
        \begin{tabular}{cccc}
            \toprule[1pt]
            \multicolumn{4}{c}{\specialcell{\bf PDVC (S3D)}}  \\
            \cmidrule(lr){1-4}
                RA & OS & SS(our) & Mix(our) \\ 
                \midrule
                5.81 & 6.00 & 6.16 & 6.55 \\
        \bottomrule[1pt]
        \end{tabular}
        \captionsetup{font=scriptsize}
        \caption{}
        \label{table:rebutal_ablation_b}
    \end{subtable}%
    \begin{subtable}{0.3\linewidth}
        \scriptsize
        \renewcommand{\tabcolsep}{0.08cm}
        \renewcommand{\arraystretch}{0.6}
        \centering
        \begin{tabular}{ccc}
            \toprule[1pt]
            \multicolumn{3}{c}{\specialcell{\bf PDVC (CLIP)}} \\
            \cmidrule(lr){1-3}
                VC & SS(our) & Mix(our) \\ 
                \midrule
                5.90 & 5.48 & 5.87 \\
        \bottomrule[1pt]
        \end{tabular}
        \captionsetup{font=scriptsize}
        \caption{}
        \label{table:rebutal_ablation_c}
    \end{subtable}%
\end{table}
\section{Additional Ablations}\label{sec:sup_ablations}
\sieveswap dataset can be scaled by varying the token-IoU ($\lambda_{IoU}$) and token-Recall ($\lambda_{Recall}$) thresholds. Fig.~\ref{fig:size-lambda}~(a) shows how varying the two $\lambda$ affects the size of the dataset. In the current work, we set thresholds based on our resources. Fig.~\ref{fig:size-lambda}~(b) shows how performance decreases when using less data. Additionally, to study the significance of individual components in our \sieveswap pipeline, we pretrain PDVC and ProcX with only sieved data and respectively finetune on YouCook2. The results in Table~\ref{table:rebutal_ablation_a} shows the significance of swapping the ASR text with human style instructions. We also mixed \sieveswap text with sieved ASR (total 1M segments) for PDVC model, obtaining SODA-C=5.87 on YouCook2 (Table~\ref{table:rebutal_ablation_c}). This boosts performance compared to only \sieveswap (SODA-C=5.48) and is comparable to pre-training on VidChapters which is 7x more data (SODA-C=5.90).
\section{Qualitative Results}\label{sec:sup_results}
\begin{figure*}
    \centering
    \captionsetup{font=scriptsize}
    \includegraphics[width=0.99\linewidth]{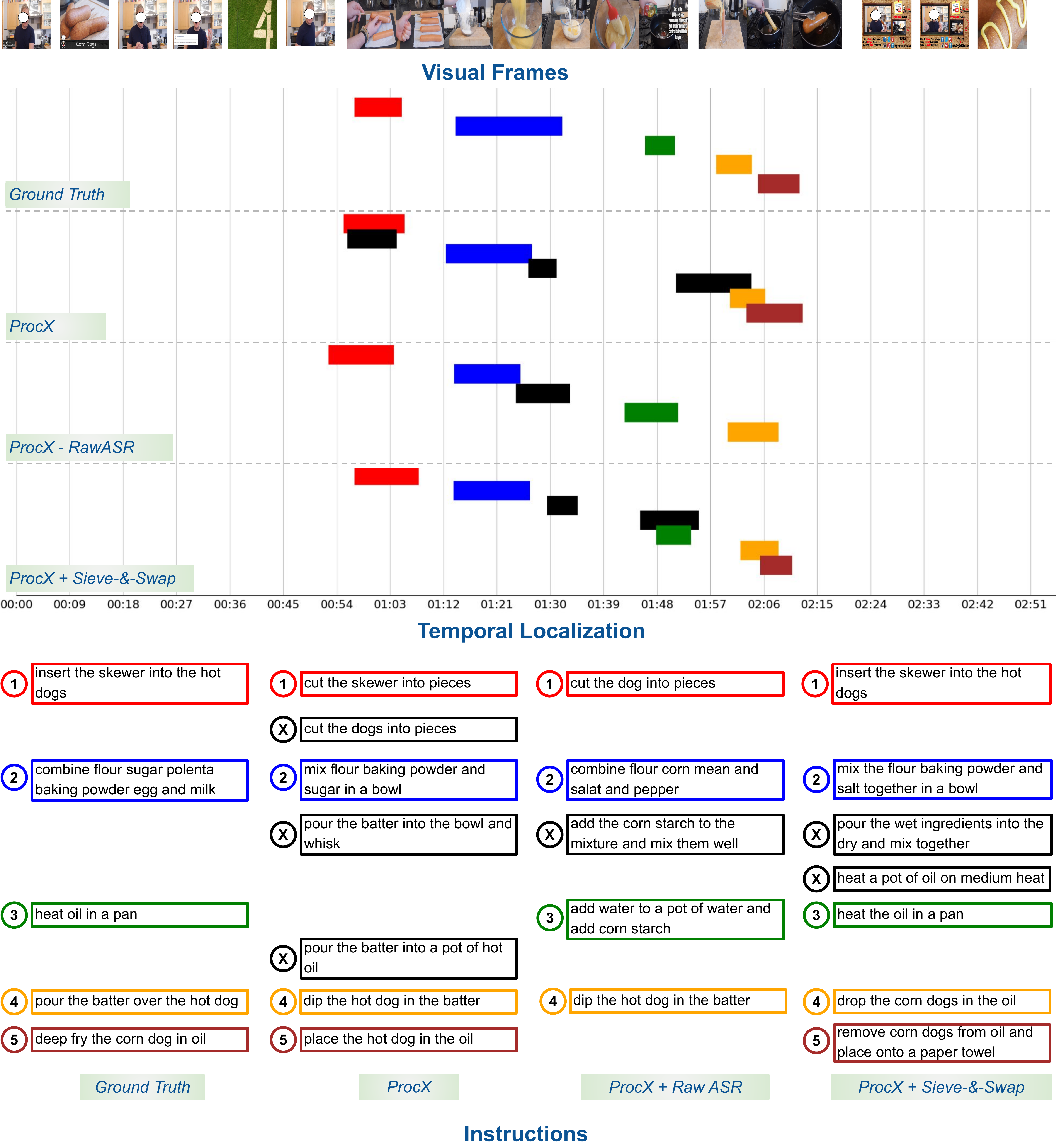}
    \caption{Qualitative Result from the test set of YouCook2 dataset with YouTube ID: `\textit{xkKuIlYSMMU}'. \textbf{Ground Truth}: The human annotated localization and written instructions. \textbf{ProcX}: The predictions are generated with ProcX (ours) architecture with no pre-training. \textbf{ProcX + Raw ASR}: The predictions are generated with ProcX, pre-trained with Raw ASR transcripts. \textbf{ProcX + \sieveswap}: The predictions are generated with ProcX, pre-trained with our \sieveswap dataset. All the architectures are fine-tuned on YouCook2. The matching segments are color-highlighted with consistent colors. Segments highlighted with black color (marked with `\textbf{X}') and missing color segments failed to align with any ground truth segments. Best visualized in color.}\label{fig:yc2_result1}
\end{figure*}

\begin{figure*}
    \centering
    \captionsetup{font=scriptsize}
    \includegraphics[width=0.99\linewidth]{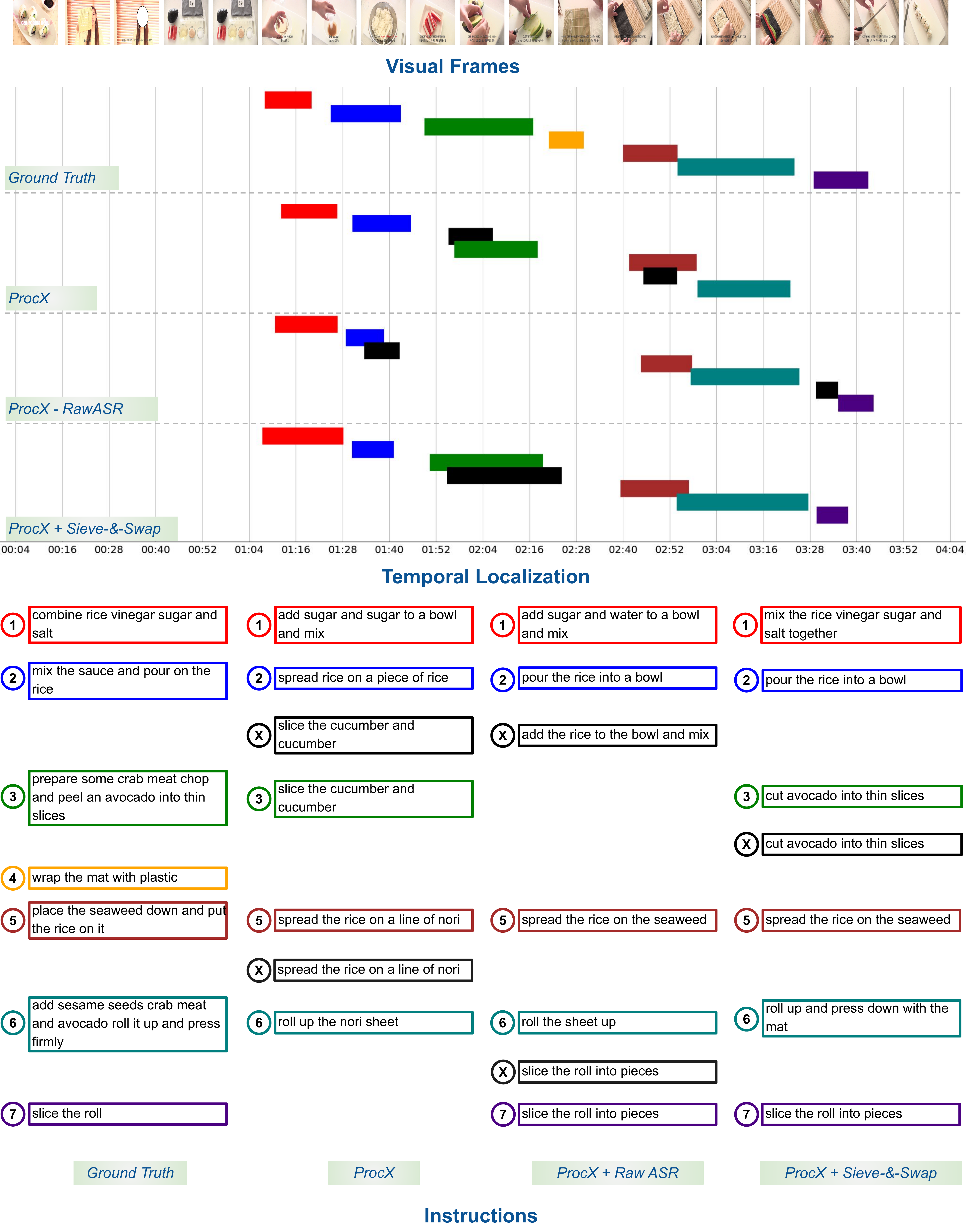}
    \caption{Qualitative Result from the test set of YouCook2 dataset with YouTube ID: `\textit{eQZEf3NCCo4}'. \textbf{Ground Truth}: The human annotated localization and written instructions. \textbf{ProcX}: The predictions are generated with ProcX (ours) architecture with no pre-training. \textbf{ProcX + Raw ASR}: The predictions are generated with ProcX, pre-trained with Raw ASR transcripts. \textbf{ProcX + \sieveswap}: The predictions are generated with ProcX, pre-trained with our \sieveswap dataset. All the architectures are fine-tuned on YouCook2. The matching segments are color-highlighted with consistent colors. Segments highlighted with black color (marked with `\textbf{X}') and missing color segments failed to align with any ground truth segments. Best visualized in color.}\label{fig:yc2_result3}
\end{figure*}

\begin{figure*}
    \centering
    \captionsetup{font=scriptsize}
    \includegraphics[width=0.99\linewidth]{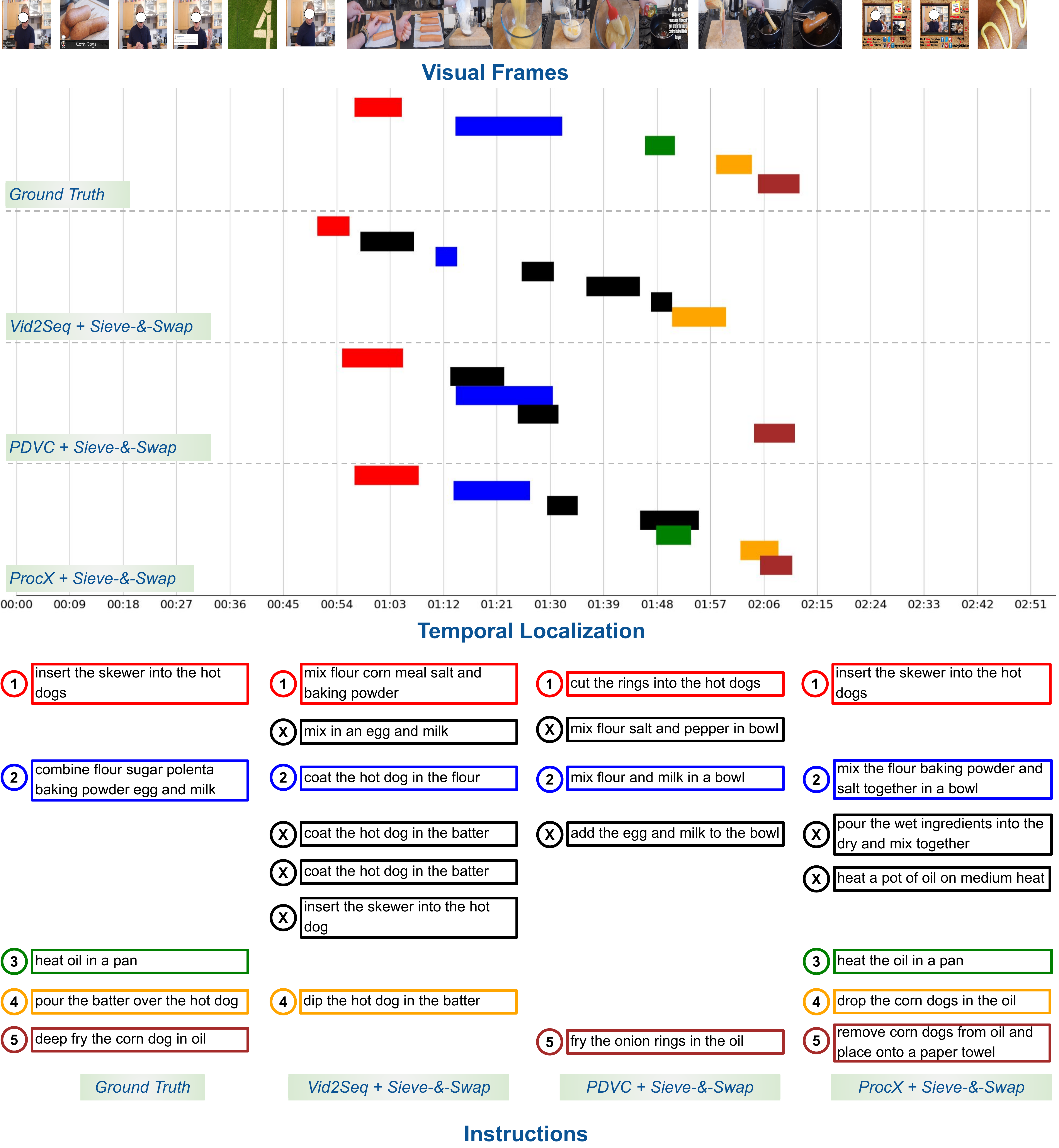}
    \caption{Qualitative Result from the test set of YouCook2 dataset with YouTube ID: `\textit{xkKuIlYSMMU}'. \textbf{Ground Truth}: The human annotated localization and written instructions. \textbf{Vid2Seq~\cite{yang2023vid2seq} + \sieveswap}: The predictions are generated with language model based Vid2Seq architecture. \textbf{PDVC~\cite{wang2021end} + \sieveswap}: The predictions are generated with LSTM based PDVC architecture. \textbf{ProcX + \sieveswap}: The predictions are generated with transformer based ProcX architecture. All the architectures are pre-trained with our \sieveswap dataset and fine-tuned on YouCook2. The matching segments are color-highlighted with consistent colors. Segments highlighted with black color (marked with `\textbf{X}') and missing color segments failed to align with any ground truth segments. Best visualized in color.}\label{fig:yc2_result2}
\end{figure*}

\begin{figure*}
    \centering
    \captionsetup{font=scriptsize}
    \includegraphics[width=0.99\linewidth]{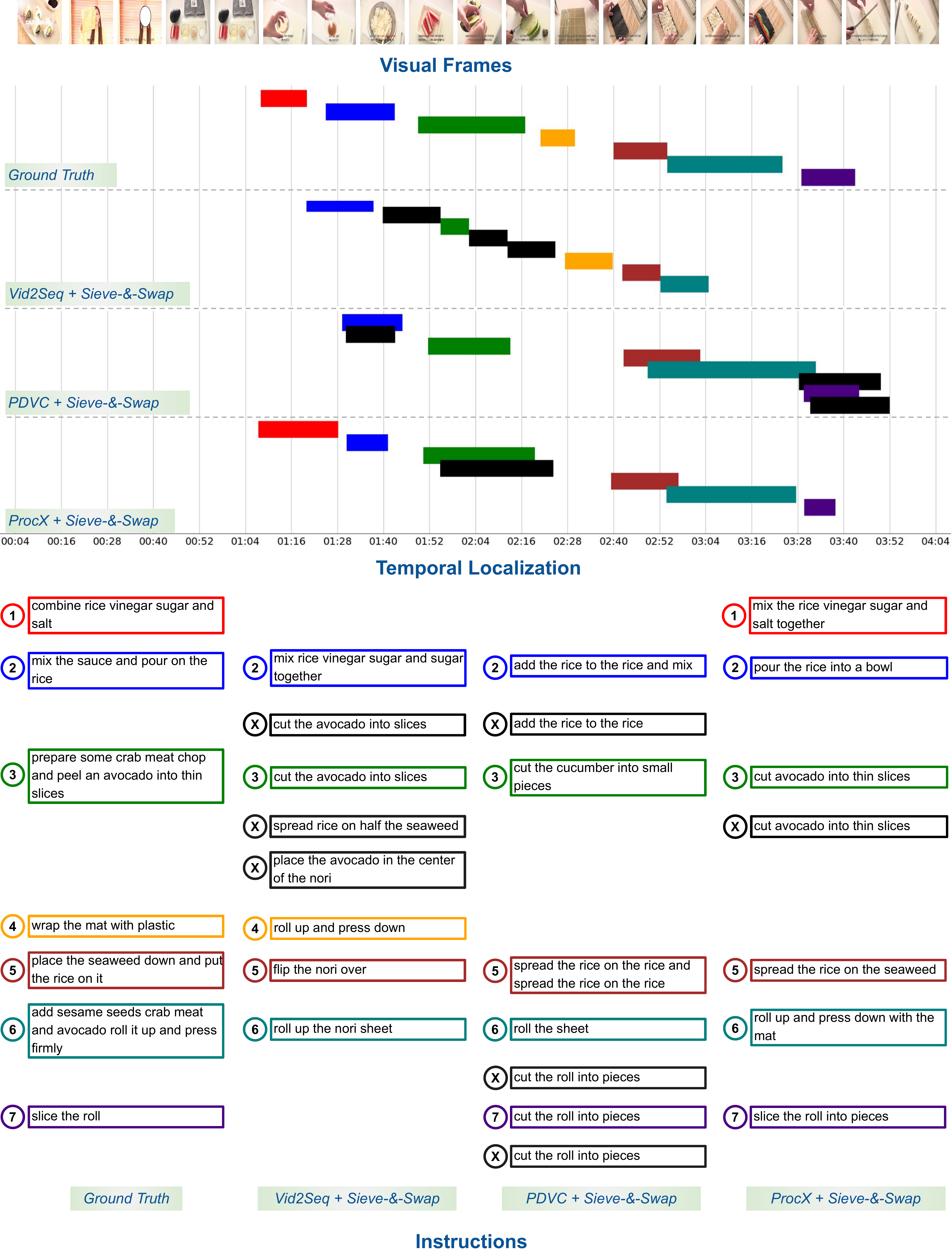}
    \caption{Qualitative Result from the test set of YouCook2 dataset with YouTube ID: `\textit{eQZEf3NCCo4}'. \textbf{Ground Truth}: The human annotated localization and written instructions. \textbf{Vid2Seq~\cite{yang2023vid2seq} + \sieveswap}: The predictions generated with language model based Vid2Seq architecture. \textbf{PDVC~\cite{wang2021end} + \sieveswap}: The predictions generated with LSTM based PDVC architecture. \textbf{ProcX + \sieveswap}: The predictions generated with transformer based ProcX architecture. All the architectures are pre-trained with our \sieveswap dataset and fine-tuned on YouCook2. The matching segments are color-highlighted with consistent colors. Segments highlighted with black color (marked with `\textbf{X}') and missing color segments failed to align with any ground truth segments. Best visualized in color.}\label{fig:yc2_result4}
\end{figure*}

\begin{figure*}
    \centering
    \includegraphics[width=0.95\linewidth]{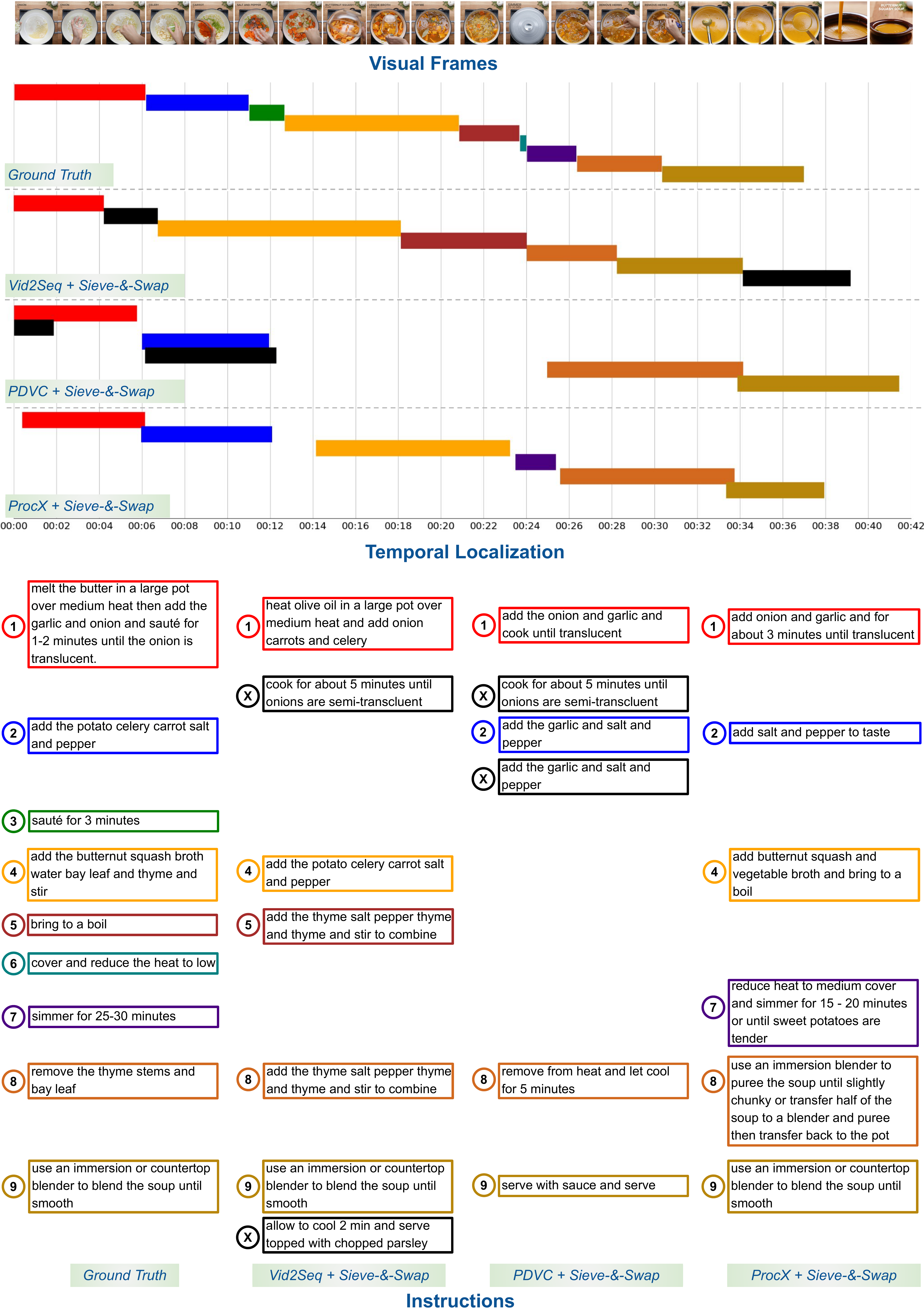}
    \captionsetup{font=scriptsize}
    \caption{Qualitative Result from the test set of Tasty dataset with dataset ID: `\textit{butternut-squash-soup}'. The human annotated localization and written instructions. \textbf{Vid2Seq~\cite{yang2023vid2seq} + \sieveswap}: The predictions are generated with language model based Vid2Seq architecture. \textbf{PDVC~\cite{wang2021end} + \sieveswap}: The predictions are generated with LSTM based PDVC architecture. \textbf{ProcX + \sieveswap}: The predictions are generated with transformer based ProcX architecture. All the architectures are pre-trained with our \sieveswap dataset and fine-tuned on Tasty. The matching segments are color-highlighted with consistent colors. Despite potential overlaps in predictions, visual clarity is ensured through the assignment of segment numbers. Segments highlighted with black color (marked with `\textbf{X}') and missing color segments are failed to align with any ground truth segments. Best visualized in color. }\label{fig:tasty_result1}
\end{figure*}

We show qualitative comparison in Figures~\ref{fig:yc2_result1},~\ref{fig:yc2_result3},~\ref{fig:yc2_result2},~\ref{fig:yc2_result4} and ~\ref{fig:tasty_result1}. Figures~\ref{fig:yc2_result1} and~\ref{fig:yc2_result3} illustrate results obtained with our ProcX model without pre-training, pre-training with raw ASR transcripts and pre-training with our \sieveswap dataset. The examples illustrate that pre-training on raw transcripts helps the model to improve the instruction coherency (SODA-C metric), however pre-training with \sieveswap achieves better coherency with less repetitiveness and redundancy. \sieveswap has marginal effect on localization improvement. This is because we do not alter the start/end times of the ASR segments, only their text content (we leave this for future work to further improve video-text alignment similar to prior work \cite{han2022temporal}). Figures~\ref{fig:yc2_result2},~\ref{fig:yc2_result4} and~\ref{fig:tasty_result1} depict results obtained pre-training Vid2Seq~\cite{yang2023vid2seq}, PDVC~\cite{wang2021end} and ProcX with \sieveswap. These examples highlight the effectiveness of the ProcX architecture with a transformer-based captioning module, as it inherits the advantages of set-based localization equipped with a transformer-based captioning module. For example, Vid2Seq~\cite{yang2023vid2seq} struggles in generating aligned instructions while predictions in PDVC~\cite{wang2021end} has more word repetition.
\begin{figure*}[t]
    \centering
    \captionsetup{font=scriptsize}
    \includegraphics[width=0.99\linewidth]{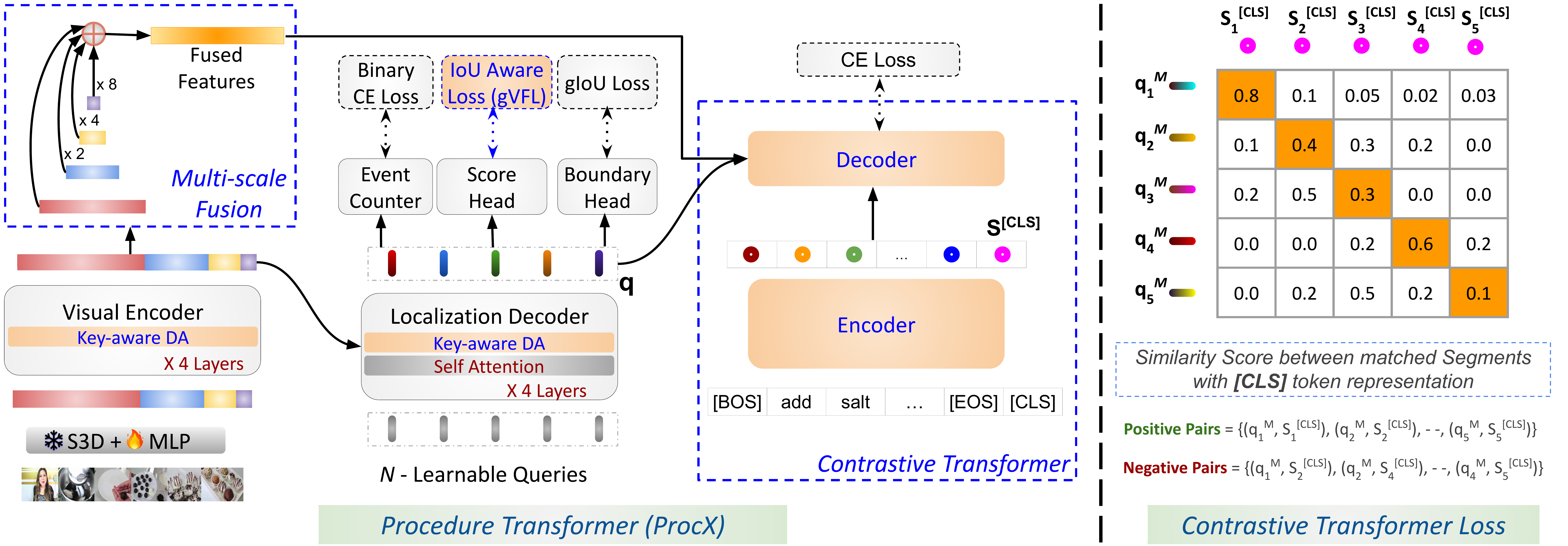}
    \caption{Our proposed ProcX architecture. The proposed components are highlighted in blue color: (i) Key-aware DA (Deformable Attention) used in visual encoder and localization decoder, (2) IoU-aware loss to better estimate the confidence score, (3) Contrastive Transformer utilizing multi-modal contrastive loss, (4) Multi-scale fusion.}
    \label{fig:procx}
\end{figure*}
\section{Architecture Details}\label{sec:sup_procx}
\myparagraph{Preliminaries - Set based Localization and Captioning.} The PDVC architecture~\cite{wang2021end} has been inspired by the DETR model~\cite{carion2020end} and formulates dense video captioning as a set prediction problem. PDVC has the following components:
\begin{enumerate}
    \item \textit{Projection Backbone}: Performs 1-D convolution with down-sampling to generate multi-scale features from pre-extracted features, e.g. S3D \cite{miech2020end}. 
    \item \textit{Visual Encoder}: A 2-layer transformer which utilizes multi-scale deformable attention~\cite{zhudeformable} to speed up the slow convergence of vanilla transformers. 
    \item \textit{$N$ Learnable Queries}: Correspond to proposal embeddings and aim to learn the representation of an instructional segment in the video.
    \item \textit{Decoder}: A 2-layer transformer with deformable cross-attention to refine the queries representation.
    \item \textit{Localization Heads}: Consists of three MLPs that predict the number of segments in a video along with a confidence score and a temporal boundary.
    \item \textit{Caption Head}: A 1-layer LSTM with deformable attention to generate the caption of each query or proposal. The LSTM layer with deformable attention performs better than a standard LSTM.
\end{enumerate}
During the training phase, PDVC utilizes the Hungarian algorithm to match proposals to the ground truth labels. During testing it utilizes the \textit{top-}$k$ proposals based on the sorted confidence score and the predicted number of segments in the video. The training objective in PDVC for the matched components is: 
\begin{equation}
\begin{split}
    \mathcal{L} = \beta_{box} \mathcal{L}_{box} + \beta_{l1} \mathcal{L}_{l1} + \beta_{cap} \mathcal{L}_{cap} \\  +  \beta_{fl} \mathcal{L}_{fl} + \beta_{count} \mathcal{L}_{count} 
\label{eq:pdvc_loss_appendix}
\end{split}
\end{equation}
where $\mathcal{L}_{cap}$ is the cross-entropy loss to predict the next token, $\mathcal{L}_{count}$ is the cross-entropy loss for segment count prediction, $\mathcal{L}_{fl}$ is the focal loss \cite{lin2017focal} between the positive and negative proposals, and $\mathcal{L}_{box}$, $\mathcal{L}_{l1}$ are the gIoU \cite{rezatofighi2019generalized} and $L1$ loss between the predicted and ground-truth temporal boundaries.

\myparagraph{Procedure Transformer (ProcX).} We follow PDVC \cite{wang2021end} and adopt the set prediction formulation due to its state-of-the-art performance. However, some design choices in PDVC are limited by the small scale of the data. For example, the neural networks are relatively shallow, being a 2-layer transformer and a single-layer LSTM. Recent work has shown significant improvements for various language and vision tasks by scaling the model size \cite{cheng2023vindlu} and utilizing transformer captioning modules trained with contrastive and generative objectives \cite{yu2022coca}. We propose to bring these advantages to PDVC, replacing the LSTM component with a transformer module to generate the captions. However, training an end-to-end transformer model is not trivial and requires a careful design with appropriate initialization \cite{popel2018training}. 

In particular, the learnable queries need to acquire a global and local representation to predict non-overlapping temporal boundaries and gain fine-grained information to generate the captions. To achieve this, we follow Contrastive Captioners (CoCa; \cite{yu2022coca}) with a modified encoder-decoder architecture trained with a contrastive and a captioning (generative) loss. The contrastive loss is responsible for learning a global representation, while the captioning loss focuses on acquiring fine-grained information. We also scale up the visual encoder and the localization decoder. The architecture of the ProcX model, along with our proposed advancements, is shown in Figure~\ref{fig:procx}. Additionally, we provide a visual representation illustrating the computation of the contrastive loss, which aims to maximize the similarity between the learnable query-based segment representation and the global $\mathrm{[CLS]}$ token, representing individual human-written text. We emphasize that the contrastive loss is calculated solely for the matched query segments. However, during the pre-training stage, it is difficult to establish a precise one-to-one correspondence between video and text due to noise in the alignment of transcripts with video content. Consequently, we incorporate the MIL-InfoNCE loss \cite{miech2020end} during pre-training and use standard InfoNCE during finetuning along with the base model objective in Equation~\ref{eq:pdvc_loss_appendix}.
\begin{align}
    L_{x_{clip}\text{-}y_{text}} & = \frac{\sum_{s \in S_+} \exp(x_s^Ty_s /\sigma)}{\sum_{s \in S_+} \exp(x_s^Ty_s / \sigma) + \sum_{s \in S_-} \exp(x_s^Ty_s / \sigma)} \\
    \mathcal{L}_{cl} & = 0.5 * (L_{x_{clip}\text{-}y_{text}} + L_{y_{text}\text{-}x_{clip}})
\end{align}
where $S_+$ and $S_-$ represent positive and negative pairs (as shown in Figure~\ref{fig:procx}), and $\sigma$ is the scaling temperature.

Contextual information is also important due to the sequential and causal nature of instructional video steps. PDVC uses deformable attention in the LSTM to sparsely attend to salient parts of the video at multiple temporal scales. However, this approach presents challenges when adapted to larger batch sizes. Consequently, we opted for a more straightforward fusion module approach to combine the multi-scale features. In particular, we employ linear interpolation to up-sample the features and subsequently sum them. The fused features are then fed to the cross-attention mechanism within the text decoder module, along with an additional learnable query. We refer to this module as \textit{Multi-Scale Fusion}, as shown in Figure~\ref{fig:procx}. This amalgamation of the aggregated multi-scale video features and the query feature effectively furnishes both local and global contextual information.
\section{Additional Details}\label{sec:sup_impl}
\subsection{Datasets}
\myparagraph{YouCook2 \cite{ZhXuCoCVPR18}.} This dataset comprises videos demonstrating 89 distinct cooking procedural tasks. The official test split of the dataset is not publicly available, hence we follow Vid2Seq~\cite{yang2023vid2seq} and report results on the validation set. Specifically, we use 840/436 training/validation annotated samples. On average, each cooking video spans about 320 seconds, and there are on average $\approx$7.7 annotated segments with a sentence per video. Like for our \sieveswap dataset, videos in YouCook2 are sourced from YouTube. To avoid an overlap between the two datasets we do not include YouTube videos annotated in YouCook2 in our \sieveswap dataset. 

\myparagraph{Tasty \cite{sener2019zero}.} This dataset is a collection of 2511 unique recipes distributed across 185 tasks, such as making cakes, pies and soups. The videos are relatively short, with an average duration of 54 seconds, and each recipe incorporates an average of nine instructions. We use the official split, utilizing 2998 videos for training, 399 for validation, and 400 for testing. Videos without transcribed speech (transcripts) present unique challenges in various scenarios, e.g., silent movies, videos created for hearing-impaired audiences, egocentric videos. The absence of transcribed speech highlights the need for models to rely solely on visual features. An illustrative example can be found on the Tasty website\footnote{https://tasty.co/}.

\subsection{Implementation Details}
In the fine-tuning stage we use the same hyperparameters used by PDVC \cite{wang2021end}: $\beta_{gIoU}=4.0$, $\beta_{fl}=2.0$, $\beta_{count}=0.5$. As our model is different from PDVC we also make a few modifications: we adjust $\beta_{l1}=0.5$, $\beta_{cap}=5.0$ and $\beta_{CL}=0.8$ for both datasets. We use AdamW \cite{loshchilov2017decoupled} optimizer with a weight decay of 0.001. During the pre-training and fine-tuning stages, we use a learning rate of 1e-4 with warm-up cosine scheduling. We use a mini-batch size of 8 for Tasty and YouCook2, while in the pre-training stage, we use a batch size of 64 (split across 4 GPUs). We pre-train the ProcX model with our \sieveswap dataset for 10 epochs and then fine-tune for 20/30 epochs for YouCook2 and Tasty datasets respectively. We pre-train Vid2Seq \cite{yang2023vid2seq} and PDVC \cite{wang2021end} for 40/30 epochs and finetune for 40/20 epochs. We clip the maximum norm of the gradient to 0.1 during the pre-training and fine-tuning stages. For regularization, we use label smoothing in the caption cross-entropy loss with a value 0.2 and dropout with a probability of 0.1. We utilize four A100-80GB GPUs for pre-training and one A100-40GB GPU for fine-tuning. The pre-training of ProcX on the \sieveswap dataset takes about $\approx 32 (8\times 4)$ GPU hours. We implement our code in PyTorch \cite{paszke2019pytorch} and will make both our data and code publicly available upon acceptance.



\end{document}